\documentclass[5p]{elsarticle}

\usepackage{listings}
\usepackage{color}
\usepackage[hidelinks]{hyperref}
\usepackage{amsmath}
\usepackage{amssymb}
\usepackage{amsthm}
\usepackage{subcaption}
\usepackage{balance}
\usepackage{stfloats}
\usepackage{graphicx}
\usepackage{epstopdf}
\usepackage{bm}
\usepackage{booktabs}
\usepackage{array}
\usepackage{stfloats}
\usepackage{pdflscape}
\usepackage{pgfplots}
\usepackage{algorithm}
\usepackage{algorithmic}
\usepackage{setspace}
\usepackage{tikz}
\usetikzlibrary{intersections}
\usetikzlibrary{angles, quotes}

\theoremstyle{plain}
\newtheorem{theorem}{Theorem}

\theoremstyle{definition}
\newtheorem{problem}{Problem}

\newtheorem{definition}{Definition}

\newtheorem{remark}{Remark}

\definecolor{mygray}{rgb}{0.5,0.5,0.5}
\definecolor{keyword}{rgb}{0.5,0.5,0.5}
\definecolor{greenCode}{rgb}{0, 0.6, 0}

\lstdefinelanguage{HTTP}{
  keywords={GET},
  ndkeywords={PUT},
  comment=[s]{PO}{T},
  morecomment=[s]{D}{LETE}
}

\lstdefinestyle{customc}{
  belowcaptionskip=1\baselineskip,
  language={HTTP},
  breaklines=true,
  frame=tb,
  captionpos=b,
  keywordstyle=\bfseries\color{greenCode},
  ndkeywordstyle=\bfseries\color{red},
  commentstyle=\bfseries\color{magenta},
  stringstyle=\bfseries\color{black},
  xleftmargin={0.75cm},
  showstringspaces=false,
  basicstyle=\footnotesize\ttfamily,
  numbers=left,
  numberstyle=\small\color{black},
}

\usepackage{flushend} 
\usepackage[colorinlistoftodos]{todonotes}

\begin{document}

\title{Correspondenceless scan--to--map-scan matching of homoriented 2D scans for mobile robot localisation}

\author[1]{Alexandros Filotheou\corref{cor1}%
\fnref{fn1}}
\ead{alexandros.filotheou@gmail.com}
\cortext[cor1]{Corresponding author}
\address[1]{Agnostou Stratiotou 10, 54631, Thessaloniki, Greece}

\thispagestyle{empty}
\pagestyle{empty}

\begin{abstract}
The objective of this study is improving the location estimate of a mobile robot
capable of motion on a plane and mounted with a conventional 2D LIDAR sensor,
given an initial guess for its location on a 2D map of its surroundings.
Documented herein is the theoretical reasoning behind solving a matching problem
between two homoriented 2D scans, one derived from the robot's physical sensor
and one derived by simulating its operation within the map, in a manner that
does not require the establishing of correspondences between their constituting
rays. Two results are proved and subsequently shown through experiments. The
first is that the true position of the sensor can be recovered with arbitrary
precision when the physical sensor reports faultless measurements and there is
no discrepancy between the environment the robot operates in and its perception
of it by the robot. The second is that when either is affected by disturbance,
the location estimate is bound in a neighbourhood of the true location whose
radius is proportional to the affecting disturbance.
\end{abstract}

\begin{keyword}
  robot localisation \sep scan--to--map-scan matching
\end{keyword}

\maketitle

\section{Introduction}
Mobile robot localisation in one plane is a well-studied field in robotics,
and several diverse approaches have been proposed in the past. Probabilistic
methods, e.g. the Kalman filter \citep{kalman}, or Monte Carlo Localisation
(MCL) methods \citep{MCL, Thrun02d}, have been applied to the task of
localisation and proved their success with respect to tracking accuracy, and
their robustness with respect to sensor noise, discrepancies between the robot's
environment and its corresponding map, motion model mismatch with regard to the
true kinematics of the robot, and pose uncertainty \citep{bible}. As for
sensors, apart from encoders, LIght Detection And Ranging devices (LIDARs) have
become popular in robot localisation due to their measurement precision,
real-time operability, and virtually no need for preprocessing.

In practice, due to an abundance of reasons---range scan measurements being
corrupted by noise, the map the robot navigates in does not match the
environment perfectly, the map is expressed as a finite resolution grid, noisy
or faulty but ever-drifting odometry---the resulting localisation estimate is
beset by an error which is often measured in centimetres or even decimetres
\citep{zhu, dm}. Apart from the conceptual challenge of reducing this
error, in certain conditions, such as industrial ones
\citep{on_the_pos_acc, paid_original}, this order of magnitude of the estimate's
error is not acceptable, and therefore prosthetic methods have been employed in
tandem with well-established sturdy probabilistic localisation methods, with
most utilising the onboard pre-existing LIDAR sensors due to their
aforementioned merits.

Such a method is introduced in \citep{paid_original}: it uses (a) MCL combined
with Kullback-Leibler Divergence (KLD) sampling \citep{KLD} as its base
probabilistic localisation method, (b) scan--to--map-scan matching for improving
the orientation estimate first, and then (c) a method that rests on the Discrete
Fourier Transform (DFT) of the two now almost homoriented scans in order to
improve the location estimate. However, the mathematical reasoning behind
the deduction of the offset estimate between the location estimate and the
robot's real location accounts only for robot orientations equal to zero, which
means that only when the robot's heading is that of the map's positive x-axis
can the location estimate be corrected. What is more, the method aiming to
recover the unknown (translation) offset is essentially a discrete-time
controller whose objective is to stabilise the robot's location estimate to the
robot's real pose, but no convergence or stability guarantees are given.

This paper focuses on the third step of the pipeline localisation method
expressed in \citep{paid_original} and, specifically, it aims to supplement it
so that it is effective over all robot orientations, by deriving the necessary
relations used in facilitating the extraction of an improved location
estimate over the entire rotation space. What is more, this paper provides
guarantees of the produced method's convergence and stability by means of
analysis resting on Lyapunovian notions of stability.

The proposed method's natural adversaries are those which can also extract the
relative translation between homoriented scans: (a) traditional, ICP-based,
scan-matching methods \citep{ICP, mbicp, plicp}, and (b) correspondenceless
and probabilistic approaches, such as the Normal Distributions Transform (NDT)
scan-matcher \citep{ndt1, ndt2, ndt3, ndt4}. However, the former are subject
to the perplexities delimited by the underlying process of establishing
correspondences between the two scans, and those posed by the plethora of
parameters governing the accuracy of their behaviour, where these must be set by
hand and address, for instance, sensor noise levels, outlier-rejection-related
variables, or particularities of the ad-hoc environment in which scan-matching
will be performed; namely: inefficient and ever-wanting tuning. Approaches
that rest on the ND transform, on the other hand, operate by discretising the
$x-y$ plane, and this fact limits their desirable accuracy and/or increases
their execution time \citep{ndt_prob1, ndt_prob2}. In contrast to ICP methods,
and similarly to the NDT method, the proposed method does not deal in
correspondences. Furthermore, it requires no parameters to be tuned at
all---apart from the maximum number of iterations, a parameter that trades
accuracy for execution time, and a numerical threshold for stopping the process
of iteration short (should the error reach a low enough level, which would make
subsequent iterations redundant). That said, the proposed method runs in real
time in modern processors. Furthermore, and most crucially, pitted against the
best-performing, state-of-the-art, ICP-based scan-matching method, and the
equally correspondenceless scan-matching method of NDT, the proposed method
achieves better accuracy and increased robustness to sensor noise and
map-to-real-world mismatch.

The remainder of the paper is structured as follows: Section
\ref{section:the_problem_and_current_solutions} formulates the problem under
purpose of solution and the solution's objective, defines necessary notions,
and finally provides a bibliographical exposition of the current
state-of-the-art solutions to the problem of performing scan--to--map-scan
matching in order to improve the pose estimate of a range-sensor-mounted robot
capable of motion in the 2D plane. Section \ref{section:the_proposed_method}
illustrates the method of solving the stated problem that this paper introduces.
The theorems on which the method rests are proved, and then its algorithmical
statement follows, accompanied with insights into its systemic operation,
convergence, and meant stability. Section \ref{section:experimental} presents
the experimental setup: a benchmark dataset is used as the sole source and
means of testing the proposed method against (a) the most accurate
correspondence-finding state-of-the-art method, and (b) the equally
corresponcenceless approach of scan-matching via NDT. Additionally, it provides
a characterisation of the proposed method in conditions arising from reality.
Finally, section \ref{section:finale} offers a recapitulation.

\section{The overall problem \& current solutions}
  \label{section:the_problem_and_current_solutions}

This section offers the formulation of the overall problem and objective aimed
at considered in this study (subsection \ref{subsec:problem_formulation}),
necessary definitions that will be useful hereafter (subsection
\ref{subsec:definitions}), and a collection of the considered problem's current
solutions (subsection \ref{subsec:sota_solutions}).

\subsection{Problem and objective formulation}
  \label{subsec:problem_formulation}

\begin{problem}
  \label{prob:the_problem}
  Let a mobile robot capable of motion in the $x-y$ plane be equipped with a
  coplanarly mounted range scan sensor, whose pose with respect to the robot's
  frame of reference is fixed, known, and of the same orientation. Let also at
  time $t \geq 0$ the following be available or standing:

  \begin{itemize}
    \item The map $M$ of the environment the robot operates in
    \item A range scan $\mathcal{S}_t^r$, captured from its range scan sensor's
          (unknown and sought for) pose $\bm{p}_t(\bm{l}_t,\theta_t)$,
          $\bm{l}_t = (x_t,y_t)$
    \item An initial estimate of the range scan sensor's pose
          $\hat{\bm{p}}_t^0(\hat{\bm{l}}_t^0, \hat{\theta}_t)$,
          where $\hat{\bm{l}}_t^0 = (\hat{x}_t^0, \hat{y}_t^0)$ is in a
          neighbourhood of $\bm{l}_t$, expressed in the map's frame of reference
    \item $|\theta_t - \hat{\theta}_t| = 0$, i.e. the real and estimated poses
          are homoriented
  \end{itemize}
\end{problem}

Then, the objective is to reduce the 2-norm of the sensor's location error
$\bm{e}(\bm{l}_t, \hat{\bm{l}}_t) \triangleq \bm{l}_t - \hat{\bm{l}}_t$ from its initial
value
\begin{align}
  \|\bm{e}(\bm{l}_t, \hat{\bm{l}}_t^0)\|_2 = ((x_t - \hat{x}_t^0)^2 + (y_t - \hat{y}_t^0)^2)^{1/2} \nonumber
\end{align}
by improving the sensor's location estimate to
$\hat{\bm{l}}_t^\prime(\hat{x}_t^\prime, \hat{y}_t^\prime)$ so that
\begin{align}
  \|\bm{e}(\bm{l}_t, \hat{\bm{l}}_t^\prime)\|_2 < \|\bm{e}(\bm{l}_t, \hat{\bm{l}}_t^0)\|_2
  \tag{$\ast$}
  \label{obj:the_objective}
\end{align}
Assuming that the sensor's pose with respect to the robot's frame of reference
is fixed at all times (as is customary in mobile robotics), this improvement in
the sensor's pose equals that of the robot's pose with respect to the map's
frame of reference.

\subsection{Definitions}
  \label{subsec:definitions}
\begin{definition}
  \label{def:definition_1}
  \textit{Definition of a range scan captured from a 2D LIDAR sensor}
  A 2D LIDAR sensor captures finitely-many ranges, i.e. distances to objects
  within its range, on a horizontal cross-section of its environment at regular
  angular and temporal intervals over a defined angular range \citep{lidar}.
  We define a range scan $\mathcal{S}$, consisting of $N$ rays over an angular
  range $\lambda$, to be an ordered sequence of $N$ pairs of (a) one
  range measurement and (b) one angle, i.e. the ray's angle relative to the
  sensor's heading, expressed in the sensor's frame of reference, ordered by
  increasing angle:
  \begin{align}
    \mathcal{S} \equiv \{(d_n, -\frac{\lambda}{2} + \frac{\lambda n}{N})\}, n = \{0,1,\dots, N-1\} \nonumber
  \end{align}
\end{definition}

\begin{figure}[H]\centering

  \begin{tikzpicture}
    \coordinate (O) at (0,0);
    \node (O_n) at (0.2,-0.2) {$O$};
    \node (x_plus) at (3.5,0) {$x$};
    \node (y_plus) at (0,3) {$y$};
    \coordinate (x_minus) at (-2,0);
    \coordinate (y_minus) at (0,-2.5);
    \coordinate (first_ray) at (-2*0.70711, -2*0.70711);
    \coordinate (first_ray_far) at (-2.5*0.70711, -2.5*0.70711);
    \node (ray_0) at (-3.0*0.70711, -3.0*0.70711){ray $0$};
    \coordinate (last_ray) at (-2*0.70711, 2*0.70711);
    \coordinate (last_ray_far) at (-2.5*0.70711, 2.5*0.70711);
    \node (ray_N) at (-3.0*0.70711, 3.0*0.70711){ray $N-1$};
    \node (l) at (-1.0,0.2){$\scriptstyle{2\pi-\lambda}$};
    \coordinate(n_c) at (3.0,1.117);
    \node[right] (n_n) at (3.0,1.117){ray $n$};
    \draw [fill] (n_c) circle [radius=0.05];
    \draw [fill] (O) circle [radius=0.05];
    \node[above] (dn) at (1.0,0.35){$d_n$};

    \draw [->] (x_minus) -- (x_plus);
    \draw [->] (y_minus) -- (y_plus);
    \draw [dashed] (O) -- (last_ray_far);
    \draw [dashed] (O) -- (first_ray_far);
    \draw [->] (O) -- (n_c);

    \draw [black, thick, dotted] (first_ray) arc[start angle=-135, end angle=135,radius=2];

    \pic [draw,  angle radius=5mm, angle eccentricity=1.4] {angle = last_ray--O--first_ray};

    \pic [draw, ->, angle radius=17mm, angle eccentricity=1.4] {angle = x_plus--O--n_c};
    \node (angle_n) at (2.6,0.44){${\scriptstyle-\dfrac{\scriptstyle\lambda}{\scriptstyle 2} + \dfrac{\scriptstyle \lambda n}{\scriptstyle N}}$};

  \end{tikzpicture}
  \caption{The (local) frame of reference of a typical range sensor. The sensor
           is located at $O(0,0)$ and its heading is that of the $x$ axis}
  \label{fig:laser}
\end{figure}
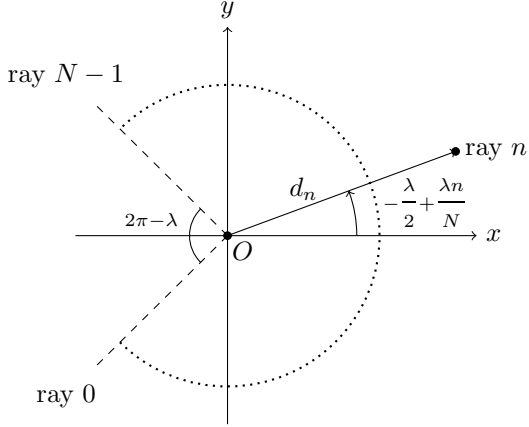

\begin{remark}
  The angular range of a LIDAR sensor is symmetrically distributed on either
  side of its $x$-axis, and each ray is equiangularly spaced from its
  neighbouring rays (with the exception of the first and last rays
  if $\lambda < 2\pi$).
\end{remark}

\begin{definition}
  \label{def:definition_2}
  \textit{Scan-to-scan matching using a 2D LIDAR sensor} (adapted for use in
  two dimensions from \citep{plicp})
  Let two range scans as defined by Definition \ref{def:definition_1},
  $\mathcal{S}_r$ and $\mathcal{S}_v$, be captured from a LIDAR
  sensor operating in the same environment at both capturing times. Let
  $\bm{p}_v(x_v,y_v,\theta_v)$ be the pose from
  which the sensor captured $\mathcal{S}_v$, expressed in some coordinate
  system (usually a past pose estimate of the sensor). The objective of
  scan-to-scan matching in two dimensions is to find the roto-translation
  $\bm{q} = (\bm{t}, \theta)$, $\bm{t} = (\Delta x, \Delta y)$ that minimises
  the distance of the endpoints of $\mathcal{S}_v$ roto-translated by
  $\bm{q}$ to their projection on $\mathcal{S}_r$. Denoting the
  endpoints of $\mathcal{S}_v$ by $\{\bm{p}_v^i\}$, in formula:
  \begin{align}
    \underset{\bm{q}}{\min} \sum\limits_i \Big\| \bm{p}_v^i \oplus \bm{q} - \prod \{ \mathcal{S}_r, \bm{p}_v^i \oplus \bm{q} \}\Big\|^2
    \label{eq:s2sm_def}
  \end{align}
  The symbol ``$\oplus$" denotes the roto-translation operator
  $\bm{p}_v^i \oplus (\bm{t}, \theta) \triangleq \bm{R}(\theta) \bm{p}^i_v + \bm{t}$,
  where $\bm{R}(\theta)$ is the 2D rotation matrix for argument angle $\theta$,
  and $\prod\{\mathcal{S}_r, \bm{p}_v^i \oplus \bm{q} \}$ denotes
  the Euclidean projector on $\mathcal{S}_r$.
\end{definition}

\begin{remark}
  The solution to (\ref{eq:s2sm_def}) cannot, in general, be found in closed
  form due to the arbitrary nature of $\mathcal{S}_r$ and the
  nonlinearity of the ``$\oplus$" operator.
\end{remark}

\begin{remark}
  Scan-to-scan matching is employed in robotics as a form and means of odometry,
  primarily in non-wheeled robots where no encoders can be utilised, or as a
  useful ameliorator of the ever-drifting encoder-ed odometry: scans captured at
  consecutive time instances, inputted to a scan-matching algorithm, convey an
  estimate as to the pose of the scan sensor at the second capture time
  relative to that captured first. It is being successfully employed in the
  tasks of Simultaneous Localisation and Mapping
  \citep{am_odom_1, am_odom_2, am_odom_3}, local map construction
  \citep{am_odom_4, am_odom_5, am_odom_6}, or in people-tracking systems
  \citep{am_odom_7}.
\end{remark}

\begin{definition}
  \label{def:definition_3}
  \textit{Definition of a map-scan}
  A map-scan is a virtual scan that encapsulates the same pieces of information
  as a scan derived from a physical sensor; only their underlying operating
  principle is different due to the fact the map-scan refers to distances to
  obstacles within the map of the robot's environment (hence its virtuality)
  rather than within the environment itself. A map-scan is captured from a
  virtual sensor whose pose relative to the robot's virtual frame of reference
  is the same as that of the physical sensor relative to the real robot's frame
  of reference, and derived by means of locating intersections of rays emanating
  from the estimate of the virtual sensor's pose with boundaries demarcating
  obstacles in the map.
\end{definition}

\begin{definition}
  \label{def:definition_4}
  \textit{Scan--to--map-scan matching in two dimensions}
  Scan--to--map-scan matching is defined in the same way as scan-to-scan
  matching but with $\mathcal{S}_v$ now derived not from the physical
  environment of the robot but from its map. This subtle difference makes
  $\bm{p}_v$, the pose from which the map-scan $\mathcal{S}_v$ was
  captured (Definition \ref{def:definition_2}), continually expressible in the
  map's frame of reference, and therefore in absolute terms, rather than
  relative to its previous estimate (recursively relative to a convention of
  the robot's starting pose).
\end{definition}

\begin{remark}
  The advantage of matching a scan derived from a physical sensor
  from its actual pose and a map-scan derived from a virtual sensor from its
  estimated pose comes now into light: Assume that robot localisation is
  performed on a mobile robot equipped with a 2D range-scan sensor via some
  localisation method which produces its pose estimate; assuming that the range
  sensor is fixed at the same pose relative to the robot in both real and
  virtual environments, the roto-translation of the virtual scan's endpoints
  that minimises their distance to their projection on the physical scan equals
  the roto-translation that, when applied to the robot's estimated pose will
  minimise its distance to its real pose. Therefore, extracting the relative
  roto-translation of the virtual scan with respect to the real scan can be
  used as a correction of the localisation estimate of the robot's pose within
  the map, thereby making the reduction of the inevitable localisation error
  inherent to localisation approaches (either probabilistic or other)
  conditionally possible.
\end{remark}

\begin{definition}
  \textit{Admissible sensor location estimates \& admissible estimate errors}
  Admissible sensor location estimates are those whose line of sight to the true
  sensor location are uninterrupted by the map---e.g. the line segment that
  connects the two does not intersect obstacles it. The estimate errors of
  admissible sensor location estimates are called admissible estimate errors.
  The term ``in a neighbourhood of the sensor's true location" or variations of
  it implies that all location estimates interior to that neighbourhood are
  admissible.
\end{definition}

\subsection{State-of-the-art approaches}
  \label{subsec:sota_solutions}
The bibliography on solving Problem \ref{prob:the_problem} is anything but
vast, and, in their majority, state-of-the-art solutions do not require that the
estimated and true poses are aligned with regard to orientation.

For example, in \citep{Chalmers} an elementary stochastic search algorithm is
employed to correct the robot's translational and rotational pose error due to
its inevitable odometric drift. This auxiliary localisation behaviour is
activated whenever an error measure that is based on the relative deviation in
detected distances between rays from a real scan and a map-scan is found to be
above a preset threshold. To avoid having to correct for the motion of the robot
while scan-matching, the robot is assumed to be standing still for the whole
duration of its pose correction. Therefore, whenever the error measure is found
to be above its preset threshold, the algorithm halts the robot's motion, and
picks a random pose in the neighbourhood of its estimated pose. It then takes a
virtual range scan from that pose, and computes the new error. If the error is
lower than the one found for the previous estimated pose, a new iteration
starts, this time centered around the newly found pose; if not, the algorithm
keeps guessing poses until it finds one whose error is lower than the previous
one. The final pose is then taken as the true pose of the robot, allowing for a
correction of the odometry. Experiments performed with this method showed that
it was able to correct a radial pose error of $0.3$ m to $0.07$ m, and an
angular pose error of $0.393$ rad to $0.01$ rad.

By contrast, in order to solve Problem \ref{prob:the_problem}, the authors of
\citep{paid_original} assume that the real and the virtual scans have already
been angularly aligned through the use of scan--to--map-scan matching, and
specifically through the all-encompassing, highly accurate, efficient, and
outperformer of the state-of-the-art scan-matchers: PL-ICP \citep{plicp}. Their
findings indicate that the improvement of the location estimate through
scan-matching is unstable, and therefore that utilising ICP-variants in order
to extract the relative translation between LIDAR-extracted scans and map-scans
is precarious and hence unsuitable in the context of localisation of autonomous
forklifts in industrial warehouse settings where milli-meter accuracy is
required. In the technical report they outline how at each localisation step
they (a) acquire a forklift's pose estimate through the use of MCL with KLD
sampling, (b) compute the relative rotation between that step's real scan and
map-scan given the forklift's pose estimate through the use of
scan--to--map-scan matching, and then, given that the forklift's orientation
error has decreased to as much as $0.13^\circ$ ($0.0023$ rad), they (c) correct
the displacement error by iteratively performing scan--to--map-scan matching
through a process that approximates the displacement error at each iteration by
a function of the first element of the Discrete Fourier transform of the
difference in ranges between that step's real scan and that iteration's
map-scan. However, in their proof of how this process is feasible and
convergent, the implication of a non-zero robot orientation is missing, which,
as we shall see in subsection \ref{subsec:method_without_disturbance} is
actually a non-trivial matter, since incorporating it leads to a result
different to that of simply rotating the location estimate vector by the
robot's orientation.

A similar pipeline is presented in \citep{gangpeng}. Instead of using PL-ICP
off-the-shelf, the authors develop a scan-matching algorithm that also aligns
real scans with map-scans taken from MCL's estimate pose using the Gauss-Newton
method during their optimisation of scan-alignment, but they do so
layer-by-layer in increasing map resolution. Experiments conducted with a real
robot in unstructured environments show that the scan--to--map-scan matcher
achieves an average location accuracy of $0.017$ m and an average rotation
accuracy of $0.0095$ rad. Interestingly, by feeding back the improved estimate
to MCL in the form of one discrete particle they manage to decrease the
location estimate error even further to achieve milli-meter accuracy.

Likewise, in \citep{hayai}, a matching algorithm that deals in range scan
features is introduced. It works by detecting rotation- and
translation-invariant features that are only computable in real-time (such as
extreme values in the polar representation of a range scan) in both real and
virtual scans before establishing correspondences between them. The
roto-translation between the two is then computed as the optimal transformation
for mapping the latter's features to the former's.

In \cite{pose_selection} scan--to--map-scan matching is employed in tandem with
a particle filter. From the pose estimate of the latter, a map-scan is computed
and then matched against the range scan captured from the physical sensor using
PLICP. Feeding back the resulting pose estimate to the population of the
particle filter in the form of a multitude of particles is shown to exhibit
lower pose errors compared to \cite{gangpeng}, where the resulting pose
estimate is fed back in the form of only one particle. Furthermore it is shown
that this method of feedback exhibits increased robustness compared to
\cite{paid_original}, where the particle filter is initialised anew around the
resulting estimate.

\section{The proposed method}
  \label{section:the_proposed_method}
We first study the unreal situation where no disturbances are acting on the
range measurements of either the real or the virtual scan. This is performed
not for theoretical reasons but because through it we establish robust stability
of the proposed solution to Problem \ref{prob:the_problem} in real conditions
where disturbances are present in both scans.

\subsection{Without disturbances}
  \label{subsec:method_without_disturbance}
\begin{theorem}
  \label{prop:theorem_without_disturbance}
  Let the assumptions of Problem \ref{prob:the_problem} hold at time $t$,
  and the angular range of the range scan sensor be $\lambda = 2\pi$.
  Let a map-scan, denoted by $\mathcal{S}_\text{v}^{M}(\hat{\bm{p}}_t)$, be
  captured from $\hat{\bm{p}}_t$ within map $M$. Assume that both
  $\mathcal{S}_\text{r}(\bm{p}_t)$ and
  $\mathcal{S}_\text{v}^{M}(\hat{\bm{p}}_t)$ range scans are disturbance-free,
  that is, the distances to obstacles the rays of the real scan capture
  correspond to the true distance of the sensor to said obstacles, and that the
  map of the environment captures the latter perfectly. Then, treating the
  estimate of the location of the sensor
  $\hat{\bm{l}}_t = [\hat{x}_t, \hat{y}_t]^{\top}$ as a state variable and
  updating it according to the difference equation
  \begin{align}
    \hat{\bm{l}}_t[k+1] = \hat{\bm{l}}_t[k] + \bm{u}[k]
    \label{eq:difference_equation_without_disturbance}
  \end{align}
  where $\hat{\bm{l}}_t[0] = \hat{\bm{l}}_t^0 = [\hat{x}_t^0, \hat{y}_t^0]^{\top}$,
  i.e. the supplied initial location estimate,
  \begin{align}
    \bm{u}[k] = \dfrac{1}{N}
    \begin{bmatrix}
      \cos\hat{\theta}_t & \sin\hat{\theta}_t \\
      \sin\hat{\theta}_t & - \cos\hat{\theta}_t
    \end{bmatrix}
    \begin{bmatrix}
      X_{1,r}\big(\mathcal{S}_\text{r}(\bm{p}_t), \mathcal{S}_\text{v}^{M}(\hat{\bm{p}}_t[k])\big) \vspace{0.2cm} \\
      X_{1,i}\big(\mathcal{S}_\text{r}(\bm{p}_t), \mathcal{S}_\text{v}^{M}(\hat{\bm{p}}_t[k])\big)
    \end{bmatrix}
    \label{eq:control_vector_without_disturbance}
  \end{align}
  is the two-dimensional vector hereafter referred to as the
  \textit{control vector},
  $X_{1,r}(\cdot)$ and $X_{1,i}(\cdot)$ are, respectively, the real and
  imaginary parts of the complex quantity $X_1$:
  \begin{align}
    X_1\big(\mathcal{S}_\text{r}(\bm{p}_t), \mathcal{S}_\text{v}^{M}(\hat{\bm{p}}_t[k])\big) =
      &X_{1,r}\big(\mathcal{S}_\text{r}(\bm{p}_t), \mathcal{S}_\text{v}^{M}(\hat{\bm{p}}_t[k])\big) \nonumber \\
      + i \cdot &X_{1,i}\big(\mathcal{S}_\text{r}(\bm{p}_t), \mathcal{S}_\text{v}^{M}(\hat{\bm{p}}_t[k])\big) \nonumber \\
      = &\sum\limits_{n=0}^{N-1}(d_n^r - d_n^v) \cdot e^{-i \frac{2 \pi n}{N}} \label{eq:X1}
  \end{align}
  where $d_n^r$ and $d_n^v$ are, respectively, the ranges of the $n$-th ray of
  the real $\mathcal{S}_\text{r}(\bm{p}_t)$ and virtual
  $\mathcal{S}_\text{v}^{M}(\hat{\bm{p}}_t[k])$ scans, and
  $\hat{\bm{p}}_t[k] = (\hat{\bm{l}}_t[k], \hat{\theta}_t)$---then
  $\hat{\bm{l}}_t[k]$ \textit{converges to} $\bm{l}_t$
  \textit{uniformly asymptotically as} $k \rightarrow \infty$.
\end{theorem}
The proof is found in subsection \ref{subsec:method_proofs_without_disturbance}.

In practice, the control system (\ref{eq:difference_equation_without_disturbance})
is let to iterate either until the norm of the control vector $\bm{u}[k]$
reaches a sufficiently small quantity: $\|\bm{u}[k]\|_2 < \varepsilon_u$, where
$\varepsilon_u$ is sufficiently small---e.g. $\varepsilon_u < 10^{-3}$,
or for $k_{max} > 0$ iterations (a sufficiently large, externally-supplied
maximum iterations threshold---e.g.  $k_{max} \geq 20$). Therefore, if
we denote by $k_{stop} \in [0, k_{max}]$ the last index of iteration, and by
$\hat{\bm{l}}_t^{\prime} = \hat{\bm{l}}_t[k_{stop}]$ $\Rightarrow$
$\|\bm{e}(\bm{l}_t, \hat{\bm{l}}_t^{\prime})\|_2 < \|\bm{e}(\bm{l}_t, \hat{\bm{l}}_t^0)\|_2$,
and therefore objective (\ref{obj:the_objective}) is attained.

\subsection{With disturbances}
  \label{subsec:method_with_disturbance}
\begin{theorem}
  \label{prop:theorem_with_disturbance}
  Let the assumptions of Problem \ref{prob:the_problem} hold at time $t$,
  and the angular range of the range scan sensor be $\lambda = 2\pi$.
  Let a map-scan, denoted by $\mathcal{S}_\text{v}^{M}(\hat{\bm{p}}_t)$, be
  captured from $\hat{\bm{p}}_t$ within map $M$. Assume that the ranges of both
  $\mathcal{S}_\text{r}(\bm{p}_t)$ and
  $\mathcal{S}_\text{v}^{M}(\hat{\bm{p}}_t)$ range scans are affected by
  additive, bounded disturbances. Then, treating the estimate of the location of
  the sensor $\hat{\bm{l}}_t = [\hat{x}_t, \hat{y}_t]^{\top}$ as a state
  variable and updating it according to the difference equation
  \begin{align}
    \hat{\bm{l}}_t[k+1] = \hat{\bm{l}}_t[k] + \widetilde{\bm{u}}[k]
    \label{eq:difference_equation_with_disturbance}
  \end{align}
  where $\hat{\bm{l}}_t[0] = \hat{\bm{l}}_t^0 = [\hat{x}_t^0, \hat{y}_t^0]^{\top}$,
  i.e. the supplied initial location estimate,
  \begin{align}
    \widetilde{\bm{u}}[k] = \dfrac{1}{N}
    \begin{bmatrix}
      \cos\hat{\theta}_t & \sin\hat{\theta}_t \\
      \sin\hat{\theta}_t & - \cos\hat{\theta}_t
    \end{bmatrix}
    \begin{bmatrix}
      \widetilde{X}_{1,r}\big(\mathcal{S}_\text{r}(\bm{p}_t), \mathcal{S}_\text{v}^{M}(\hat{\bm{p}}_t[k])\big) \vspace{0.2cm} \\
      \widetilde{X}_{1,i}\big(\mathcal{S}_\text{r}(\bm{p}_t), \mathcal{S}_\text{v}^{M}(\hat{\bm{p}}_t[k])\big)
    \end{bmatrix}
    \label{eq:control_vector_with_disturbance}
  \end{align}
  $\widetilde{X}_{1,r}(\cdot)$ and $\widetilde{X}_{1,i}(\cdot)$ are,
  respectively, the real and imaginary parts of the complex quantity
  $\widetilde{X}_1$:
  \begin{align}
    \widetilde{X}_1\big(\mathcal{S}_\text{r}(\bm{p}_t), \mathcal{S}_\text{v}^{M}(\hat{\bm{p}}_t[k])\big) =
      &\widetilde{X}_{1,r}\big(\mathcal{S}_\text{r}(\bm{p}_t), \mathcal{S}_\text{v}^{M}(\hat{\bm{p}}_t[k])\big) \nonumber \\
      + i \cdot &\widetilde{X}_{1,i}\big(\mathcal{S}_\text{r}(\bm{p}_t), \mathcal{S}_\text{v}^{M}(\hat{\bm{p}}_t[k])\big) \nonumber \\
      = &\sum\limits_{n=0}^{N-1}(\widetilde{d}_n^r - \widetilde{d}_n^v) \cdot e^{-i \frac{2 \pi n}{N}} \nonumber 
  \end{align}
  where $\widetilde{d}_n^r$ and $\widetilde{d}_n^v$ are, respectively, the
  perturbed ranges of the $n$-th ray of the real
  $\mathcal{S}_\text{r}(\bm{p}_t)$ and virtual
  $\mathcal{S}_\text{v}^{M}(\hat{\bm{p}}_t[k])$ scans, and
  $\hat{\bm{p}}_t[k] = (\hat{\bm{l}}_t[k], \hat{\theta}_t)$---then
  $\hat{\bm{l}}_t[k]$ is uniformly bounded for $k \geq k_0$ and uniformly
  ultimately bounded in a neighbourhood of $\bm{l}_t$ whose size depends on the
  suprema of the disturbance corrupting the range measurements of the two
  scans.
\end{theorem}
The proof is found in subsection \ref{subsec:method_proofs_with_disturbance}.

The above theorem provides guarantees of the proposed method's convergence in
real conditions, where the measurements of a physical 2D LIDAR sensor are
accurate to a certain extent, and where the same applies to the ranges reported
by the virtual range sensor due to the imperfection with which the map of the
environment in which a sensor is placed is able to capture its object.

Compared to the case where no disturbances are present, a solution
satisfying objective (\ref{obj:the_objective}) is not strictly
guaranteed for all admissible $\hat{\bm{l}}^0_t$. Let us again denote by
$k_{stop} \in [0, k_{max}]$ the last index of iteration, by
$\hat{\bm{l}}_t^{\prime} = \hat{\bm{l}}_t[k_{stop}]$ the final estimate of the
sensor's location, and by $B$ the ultimate bound of the error. If
$\|\bm{e}(\bm{l}_t, \hat{\bm{l}}_t^0)\|_2 > B$, Theorem
\ref{prop:theorem_with_disturbance} guarantees the satisfaction of objective
(\ref{obj:the_objective}) if $k_{stop} \geq k_0$. If, on the other hand,
$\|\bm{e}(\bm{l}_t, \hat{\bm{l}}_t^0)\|_2 \leq B$, it is not certain that
$\|\bm{e}(\bm{l}_t, \hat{\bm{l}}_t^{\prime})\|_2 < \|\bm{e}(\bm{l}_t, \hat{\bm{l}}_t^0)\|$;
what is certain in this case, though, is that $\|\bm{e}(\bm{l}_t, \hat{\bm{l}}_t[k])\|_2 \ngtr B$
for all $k \geq 0$.

\subsection{Proof of convergence and stability when disturbances are absent}
  \label{subsec:method_proofs_without_disturbance}
For readability of the below proof purposes we shall hereafter in this subsection
adopt a friendlier notation than that used in subsection
\ref{subsec:method_without_disturbance}.

Let a range scan $\mathcal{S}$, consisting of $N$ rays over an angular range of
$2\pi$, be represented by an ordered sequence of $N$ pairs of (a) one range
measurement and (b) one angle, i.e. the ray's angle relative to the range
sensor's heading, ordered by increasing angle:
$\mathcal{S} \equiv \{(d_n,-\pi + \frac{2 \pi n}{N})\}$, $n = \{0,1,\dots, N-1\}$.
Assuming that relative to the map the sensor is located at $(x_s,y_s)$ and that
its orientation relative to the $x$-axis of the map' s frame of reference is
$\theta \in \mathbb{R}$ and known, the coordinates of the end-point of the
scan's $n$-th ray within the map's frame of reference are $(x_n,y_n)$, where:
\begin{align}
  x_n -x_s &= d_n cos(\theta + \frac{2 \pi n}{N} - \pi) = -d_n cos(\theta + \frac{2 \pi n}{N}) \label{eq:x_n}\\
  y_n -y_s &= d_n sin(\theta + \frac{2 \pi n}{N} - \pi) = -d_n sin(\theta + \frac{2 \pi n}{N}) \label{eq:y_n}
\end{align}
Here we make the observation that $-(x_n-x_s)$ and $y_n-y_s$ are, respectively,
the real and imaginary parts of the complex quantity
\begin{align}
  d_n e^{-i(\theta + \frac{2 \pi n}{N})} &= d_n cos(\theta + \frac{2 \pi n}{N}) - i \cdot d_n sin(\theta + \frac{2 \pi n}{N}) \nonumber \\
                                         &\stackrel{(\ref{eq:x_n}),(\ref{eq:y_n})}{=} -(x_n-x_s) + i \cdot (y_n-y_s) \label{eq:dne_complex_x_y}
\end{align}
and, therefore, that
\begin{align}
  d_n e^{-i \frac{2 \pi n}{N}} &= e^{i\theta}(-(x_n-x_s) + i \cdot (y_n-y_s)) \label{eq:dne_complex}
\end{align}
Ergo, denoting with the superscript $r$ quantities which correspond to the
real scan $\mathcal{S}_r$, which has been captured from the unknown sensor pose
$\bm{p}(x_s^r,y_s^r,\theta_s^r)$, and with $v$ those which correspond to the virtual
scan $\mathcal{S}_v$, which has been captured from pose $\hat{\bm{p}}(x_s^v,y_s^v,\theta_s^v)$,
where $\theta_s^r = \theta_s^v = \theta$:
\begin{align}
  d_n^r e^{-i \frac{2 \pi n}{N}} &= e^{i\theta}(-(x_n^r-x_s^r) + i \cdot (y_n^r-y_s^r)) \label{eq:dne_complex_r} \\
  d_n^v e^{-i \frac{2 \pi n}{N}} &= e^{i\theta}(-(x_n^v-x_s^v) + i \cdot (y_n^v-y_s^v)) \label{eq:dne_complex_v}
\end{align}

The first term of the Discrete Fourier Transform of the signal that consists of
the difference of the two signals (\ref{eq:dne_complex_r}) and
(\ref{eq:dne_complex_v}) is
\begin{align}
  X_1 &= \sum\limits_{n=0}^{N-1}(d_n^r - d_n^v) \cdot e^{-i \frac{2 \pi n}{N}} \nonumber \\
      &=\sum\limits_{n=0}^{N-1}(d_n^r e^{-i \frac{2 \pi n}{N}} - d_n^v e^{-i \frac{2 \pi n}{N}}) \nonumber \\
      &\stackrel{(\ref{eq:dne_complex_r}),(\ref{eq:dne_complex_v})}{=}  \sum\limits_{n=0}^{N-1}(e^{i\theta}(-(x_n^r-x_s^r) + i \cdot (y_n^r-y_s^r)) \nonumber\\
      &\ \ \ \ \ \ \ \ \ \ \ - e^{i\theta}(-(x_n^v-x_s^v) + i \cdot (y_n^v-y_s^v))) \nonumber \\
      &=e^{i\theta}\sum\limits_{n=0}^{N-1}((-(x_n^r-x_s^r) + i \cdot (y_n^r-y_s^r)) \nonumber \\
      &\ \ \ \ \ \ \ \ \ \ \ -(-(x_n^v-x_s^v) + i \cdot (y_n^v-y_s^v))) \nonumber \\
      &=e^{i\theta}(-\sum\limits_{n=0}^{N-1}(x_n^r - x_n^v) + i \cdot \sum\limits_{n=0}^{N-1}(y_n^r - y_n^v)) \nonumber \\
      &+e^{i\theta}(\sum\limits_{n=0}^{N-1}(x_s^r - x_s^v) - i \cdot \sum\limits_{n=0}^{N-1}(y_s^r - y_s^v)) \label{eq:X1_before_approx}
\end{align}
In the first summand of (\ref{eq:X1_before_approx}), the quantities under
summation express the $x$-wise and $y$-wise offsets of the endpoint of
the virtual scan's $n$-th ray from the real scan's $n$-th ray, expressed in the
map's frame of reference. In general, they cannot be calculated in closed form
since the sensor's true location is unknown, and the environment---even if it is
of absolute fidelity to the map---is arbitrary. For posterior convenience we
denote the first summand of (\ref{eq:X1_before_approx}) by
\begin{align}
  e^{i\theta}(-\sum\limits_{n=0}^{N-1}(x_n^r - x_n^v) &+ i \cdot \sum\limits_{n=0}^{N-1}(y_n^r - y_n^v)) \nonumber \\
                                                                & = N e^{i\theta}(-\delta_x + i \cdot \delta_y) \label{eq:first_summand}
\end{align}
where
\begin{align}
  \delta_x = \dfrac{1}{N} \sum\limits_{n=0}^{N-1}(x_n^r - x_n^v) \label{eq:delta_x} \\
  \delta_y = \dfrac{1}{N} \sum\limits_{n=0}^{N-1}(y_n^r - y_n^v) \label{eq:delta_x}
\end{align}
As regards the second summand of (\ref{eq:X1_before_approx}), let us denote
the components of the error between the true location $(x_s^r,y_s^r)$ and the
estimated one $(x_s^v, y_s^v)$ by $e_x = x_s^r - x_s^v$ and
$e_y = y_s^r - y_s^v$. The quantities under summation in the second summand of
the right-hand side of (\ref{eq:X1_before_approx}) do not depend on $n$, and
therefore
\begin{align}
  \sum\limits_{n=0}^{N-1}(x_s^r - x_s^v) &= \sum\limits_{n=0}^{N-1}e_x = N \cdot e_x  \label{eq:x_approx} \\
  \sum\limits_{n=0}^{N-1}(y_s^r - y_s^v) &= \sum\limits_{n=0}^{N-1}e_y = N \cdot e_y \label{eq:y_approx}
\end{align}
Then, equation (\ref{eq:X1_before_approx}) is transformed as follows:
\begin{align}
  X_1 \stackrel{(\ref{eq:first_summand}),(\ref{eq:x_approx}),(\ref{eq:y_approx})}{=} N e^{i\theta} (e_x - \delta_x -i \cdot (e_y - \delta_y))
  \label{eq:X1_after_approx}
\end{align}
Since $X_1$ is, in general, complex, it can be written in the form
$X_1 = X_{1,r} + i \cdot X_{1,i}$, where $X_{1,r}$ and $X_{1,i}$ are known
quantities ($X_1$ is the first term of the DFT of the difference between
sequences of known real numbers). Then, from equation
(\ref{eq:X1_after_approx}), the following equality is established:
\begin{align}
  X_1 &= N e^{i\theta} (e_x - \delta_x -i \cdot (e_y-\delta_y)) \nonumber \\
      &= X_{1,r} + i \cdot X_{1,i} \nonumber
\end{align}
from which the expressions for the positional errors between the estimated
and true location of the sensor $e_x$ and $e_y$ can be derived:
\begin{align}
  \label{eq:final_errors_xy}
  \begin{bmatrix}
    e_x \\
    e_y
  \end{bmatrix} &= \dfrac{1}{N}
  \begin{bmatrix}
    \cos\theta & \sin\theta \\
    \sin\theta & - \cos\theta
  \end{bmatrix}
  \begin{bmatrix}
    X_{1,r} \\
    X_{1,i}
  \end{bmatrix}+
  \begin{bmatrix}
    \delta_x \\
    \delta_y
  \end{bmatrix} \nonumber \\ \nonumber \\
  &=
  \begin{bmatrix}
    u_x \\
    u_y
  \end{bmatrix}+
  \begin{bmatrix}
    \delta_x \\
    \delta_y
  \end{bmatrix}
\end{align}
and this is the positional error between the true and estimated poses of the
sensor. Since
\begin{align}
  \begin{bmatrix}
    e_x \\
    e_y
  \end{bmatrix} =
  \begin{bmatrix}
    x_s^r \\
    y_s^r
  \end{bmatrix} -
  \begin{bmatrix}
    x_s^v \\
    y_s^v
  \end{bmatrix} \nonumber
\end{align}
and therefore
\begin{align}
  \begin{bmatrix}
    x_s^v \\
      y_s^v
  \end{bmatrix}
  +
  \begin{bmatrix}
    e_x \\
    e_y
  \end{bmatrix} =
  \begin{bmatrix}
    x_s^r \\
    y_s^r
  \end{bmatrix} \nonumber
\end{align}
adding (\ref{eq:final_errors_xy}) to the estimated position of the sensor
would allow us to extract its true position in one step. However, although the
first summand of the right-hand side of (\ref{eq:final_errors_xy}) does consist
of known quantities, the second one does not, and therefore it is impossible to
extract the sensor's true position in one step. This is the reason an iterative
process of updating the sensor's location estimate is necessary.

Let us now denote by $\bm{\hat{l}}[k] = [x_s^v[k], y_s^v[k]]^{\top}$ the
estimated location of the range sensor at iteration $k$, and by
$\bm{u}[k] = [u_x[k], u_y[k]]^{\top}$. The claim of Theorem
\ref{prop:theorem_without_disturbance} is that by updating the location estimate
with
\begin{align}
  \bm{\hat{l}}[k+1] = \bm{\hat{l}}[k] + \bm{u}[k]
  \label{eq:proof_unperturbed_system}
\end{align}
when $\bm{\hat{l}}[0] = [x_s^v[0], y_s^v[0]]^{\top}$ is set to the initially
supplied location estimate, the location estimate converges asymptotically to
the sensor's true pose $[x_s^r,y_s^r]^{\top}$ as $k \rightarrow \infty$.

In order to investigate the convergence and stability of
(\ref{eq:proof_unperturbed_system}), we first rewrite it so that it reflects the
dynamics of the error between the estimated range sensor location and the true
one. We denote this error by $\bm{e} = [e_x, e_y]^{\top}$. Then, by simply
multiplying both sides of (\ref{eq:proof_unperturbed_system}) with $-1$ and
adding $[x_s^r,y_s^r]^{\top}$ to both sides, (\ref{eq:proof_unperturbed_system})
is transformed to:
\begin{align}
  \bm{e}[k+1] = \bm{e}[k] - \bm{u}[k]
  \label{eq:proof_unperturbed_system_error}
\end{align}
We shall begin by examining the $x$-wise component of the error (the
analysis regarding the $y$-wise component is analogous) and state the two main
equations available:
\begin{align}
  e_x[k+1] &= e_x[k] - u_x[k] \label{eq:diff_eq_x} \\
  e_x[k] &= u_x[k] + \delta_x[k] \label{eq:diff_eq_x2}
\end{align}
from which we obtain that $\delta_x[k] = e_x[k+1]$.

When the estimated location of the sensor is in a neighbourhood of its true
location and space is sampled sufficiently densely ($N \gg 4$), $\delta_x$ is
strictly increasing with respect to the $x$-wise location error between
subsequent iterations:
\begin{align}
  \dfrac{\delta_x[k+1] - \delta_x[k]}{e_x[k+1] - e_x[k]} &> 0 \label{eq:delta_inc_x}
\end{align}

Our first point shall be to prove that the derivative of $u_x$ with respect to
the error $e_x$ is bounded above. From (\ref{eq:diff_eq_x2}) for $k \leftarrow k+1$:
\begin{align}
  e_x[k+1] &= u_x[k+1] + \delta_x[k+1] \nonumber \\ \noalign{\vskip2pt}
  e_x[k+1] - e_x[k] &= u_x[k+1] - e_x[k] + \delta_x[k+1] \nonumber \\ \noalign{\vskip2pt}
  \delta_x[k+1] &= (e_x[k] - u_x[k+1]) \nonumber \\ \noalign{\vskip2pt}
                &+ (e_x[k+1] - e_x[k]) \nonumber \\ \noalign{\vskip2pt}
  \delta_x[k+1] - \delta_x[k] &= (e_x[k] -\delta_x[k] - u_x[k+1]) \nonumber \\
                              &+ (e_x[k+1] - e_x[k]) \nonumber \stackrel{(\ref{eq:diff_eq_x2})}{\Longleftrightarrow}\\ \noalign{\vskip2pt}
  \delta_x[k+1] - \delta_x[k] &= (u_x[k] - u_x[k+1]) \nonumber \\ \noalign{\vskip2pt}
                              &+ (e_x[k+1] - e_x[k]) \nonumber \\ \noalign{\vskip2pt}
  \dfrac{\delta_x[k+1] - \delta_x[k]}{e_x[k+1] - e_x[k]} &= \dfrac{u_x[k] - u_x[k+1]}{e_x[k+1] - e_x[k]} + 1 \stackrel{(\ref{eq:delta_inc_x})}{>} 0 \nonumber
\end{align}

Therefore
\begin{align}
  \dfrac{u_x[k+1] - u_x[k]}{e_x[k+1] - e_x[k]} < 1 \label{eq:g1}
\end{align}

Given this, we shall prove that the derivative of $\delta_x$ with respect to
the input $u_x$ is positive. From (\ref{eq:diff_eq_x2}) for $k \leftarrow k+1$:

\begin{align}
  e_x[k+1] &= u_x[k+1] + \delta_x[k+1] \nonumber \\ \noalign{\vskip2pt}
  \delta_x[k] &= u_x[k+1] + \delta_x[k+1] \nonumber \\ \noalign{\vskip2pt}
  \delta_x[k+1] - \delta_x[k] &= -u_x[k+1] \nonumber \\ \noalign{\vskip2pt}
  \delta_x[k+1] - \delta_x[k] &= -u_x[k+1] + u_x[k] - u_x[k] \nonumber \\ \noalign{\vskip2pt}
  \dfrac{\delta_x[k+1] - \delta_x[k]}{u_x[k+1] - u_x[k]} &= -1 - \dfrac{u_x[k]}{u_x[k+1] - u_x[k]} \nonumber \\ \noalign{\vskip2pt}
  &\stackrel{(\ref{eq:diff_eq_x})}{=} -1 - \dfrac{e_x[k] - e_x[k+1]}{u_x[k+1] - u_x[k]} \nonumber \\ \noalign{\vskip2pt}
  &= -1 + \dfrac{e_x[k+1] - e_x[k]}{u_x[k+1] - u_x[k]} \nonumber \\ \noalign{\vskip2pt}
  &\stackrel{(\ref{eq:g1})}{>} 0 \nonumber
\end{align}

Now, given from (\ref{eq:delta_x}) that $\delta_x = 0$ when $e_x = 0$
(when the virtual sensor is posed at the pose of the true sensor, both
the virtual and real scan perceive exactly the same points in space, and
therefore $x_n^r = x_n^v$ for all $n \in [0, N)$), and that $u_x = 0$ when
$e_x = 0$ (by the same reasoning, from the definition of $X_1$, $X_1 = 0$
because $d_n^r = d_n^v$ for all $n \in [0, N)$), we conclude that $\delta_x = 0$
when $u_x = 0$. Since $\delta_x$ increases as $u_x$ increases, \textit{we
conclude that the input $u_x$ exhibits the same sign as $\delta_x$}. Hence,
from (\ref{eq:diff_eq_x2}) we conclude that $e_x, u_x, \delta_x$ all exhibit
the same sign, and therefore that $|u_x| \leq |e_x|$ and $|\delta_x| \leq |e_x|$
for all admissible $e_x$.

Let now $V_x : \mathbb{R} \rightarrow \mathbb{R}$ be $V_x(e_x) = e_x^2$,
where $\alpha_x(|e_x|) \leq V_x(e_x) \leq \beta_x(|e_x|)$, with $\alpha_x(\cdot)$,
$\beta_x(\cdot)$ class $\mathcal{K}$ functions \citep{khalil} for all admissible
$e_x$. It is clear that $V_x$ is a $C^1$ function, $V_x(0) = 0$, and positive
everywhere else. Then
\begin{align}
  \Delta V_x &= V_x(e_x[k+1]) - V_x(e_x[k]) \nonumber \\
           &\stackrel{(\ref{eq:diff_eq_x2})}{=} \delta_x^2[k] - (\delta_x[k] + u_x[k])^2 \label{eq:dv_plain} \\
           & \leq 0 \nonumber
\end{align}
because $|\delta_x + u_x| \geq |\delta_x|$ since $\delta_x$ and $u_x$ are of the
same sign. We now rewrite (\ref{eq:dv_plain}):
\begin{align}
  \Delta V_x &= V_x(e_x[k+1]) - V_x(e_x[k]) \nonumber \\
           &= \delta_x^2[k] - e_x^2[k] \nonumber \\
           &\stackrel{(\ref{eq:diff_eq_x})}{=} -e_x^2[k] + (e_x[k] - u_x[k])^2 \label{eq:base_for_breaking} \\
           &= u_x[k](u_x[k] - 2e_x[k]) \label{eq:breaking_eq}
\end{align}

Since $\Delta V_x \leq 0$, expression (\ref{eq:breaking_eq}) is nonpositive, and
therefore, omitting brackets since all variables refer to iteration $k$, if
$u_x \geq 0 \Rightarrow 0 \leq u_x \leq 2e_x$, since $e_x$ and $u_x$ share the
same sign. Alternatively, if $u_x \leq 0 \Rightarrow 0 \leq 2e_x \leq u_x$ for
the same reason. Now consider the former case:
\begin{align}
  0 \leq &u_x \leq 2e_x \nonumber \\
  -e_x \leq &u_x-e_x \leq e_x \nonumber
\end{align}
Since $|u_x| \leq |e_x|$ and both $u_x$ and $e_x$ are positive: $u_x \leq e_x$,
which means that $u_x - e_x \leq 0$. Then, multiplying all sides with
$(u_x - e_x)$:
\begin{align}
  -e_x \leq &u_x-e_x \leq e_x \nonumber \\
  -e_x(u_x - e_x) \geq &(u_x-e_x)^2 \geq e_x (u_x - e_x) \nonumber \\
  e_x(u_x - e_x) \leq &(u_x-e_x)^2 \leq -e_x (u_x - e_x) \nonumber
\end{align}
Therefore, from (\ref{eq:base_for_breaking}):
\begin{align}
  \Delta V_x &= -e_x^2 + (e_x - u_x)^2 \nonumber \\
           &\leq -e_x^2 -e_x (u_x - e_x) \nonumber \\
           &= - u_x e_x \nonumber
\end{align}
Now consider the latter case:
\begin{align}
  2e_x \leq &u_x \leq 0 \nonumber \\
  e_x \leq &u_x - e_x \leq -e_x \nonumber
\end{align}
Since $|u_x| \leq |e_x|$ and both $u_x$ and $e_x$ are negative: $u_x \geq e_x$,
which means that $u_x - e_x \geq 0$. Then, multiplying all sides with
$(u_x - e_x)$:
\begin{align}
  e_x \leq &u_x - e_x \leq -e_x \nonumber \\
  e_x (u_x - e_x) \leq &(u_x - e_x)^2 \leq -e_x  (u_x - e_x) \nonumber
\end{align}
Therefore, from (\ref{eq:base_for_breaking}):
\begin{align}
  \Delta V_x &= -e_x^2 + (e_x - u_x)^2 \nonumber \\
           &\leq -e_x^2 -e_x (u_x - e_x) \label{eq:shrinked} \\
           &= - u_x e_x \nonumber
\end{align}
Therefore, for all admissible $e_x$: $\Delta V_x \leq -u_x e_x$. The proof is
analogous for the $y$-wise component, and hence omitted.

Let now $V: \mathbb{R}^2 \rightarrow \mathbb{R}$ be $V(\bm{e}) = \|\bm{e}\|_2^2$,
where $\alpha(\|\bm{e}\|) \leq V(\bm{e}) \leq \beta(\|\bm{e}\|)$, with
$\alpha(\cdot)$, $\beta(\cdot)$ class $\mathcal{K}$ functions for all admissible
$\bm{e}$. It is clear that $V$ is a $C^1$ function, $V(\bm{0}) = 0$, and
positive everywhere else. Then
\begin{align}
  \Delta V &= V(\bm{e}[k+1]) - V(\bm{e}[k]) \nonumber \\
           &= \Bigg\|
           \begin{bmatrix}
             \delta_x[k] \\
             \delta_y[k]
           \end{bmatrix}
           \Bigg\|_2^2
           -
           \Bigg\|
           \begin{bmatrix}
             e_x[k] \\
             e_y[k]
           \end{bmatrix}
           \Bigg\|_2^2 \nonumber \nonumber
\end{align}
Dropping brackets for readability since all terms refer to the same iteration:
\begin{align}
  \Delta V &= (\delta_x^2 - e_x^2) + (\delta_y^2 - e_y^2) \nonumber \\
           &= - e_x^2 + (e_x - u_x)^2 -e_y^2 + (e_y - u_y)^2 \nonumber \\
           &\stackrel{(\ref{eq:shrinked})}{\leq} -e_x^2 -e_x(u_x - e_x) -e_y^2 -e_y(u_y - e_y) \nonumber \\
           &= -u_x e_x - u_y e_y \nonumber \\
           &= - \bm{u}^{\top} \bm{e} \nonumber \\
           &= -\gamma(\|\bm{e}\|_2) \nonumber
\end{align}
where $\gamma(\cdot)$ is a class $\mathcal{K}$ function. The latter equation
stands because $\bm{u}^{\top}\bm{e} = \|\bm{u}\| \|\bm{e}\| \cos\phi \geq 0$,
where $\phi$ is the angle between vectors $\bm{u}$ and $\bm{e}$, which, because
$u_x e_x \geq 0$ and $u_y e_y \geq 0$, is acute: $\phi \in [0, \pi/2]$, and
therefore $\cos\phi \geq 0$. Hence, from \citep{stability} (Theorem $2.3$ p.$26$,
points A$2$ p.$27$ and B$1$ p.28) the origin of the system
(\ref{eq:proof_unperturbed_system_error}) is uniformly asymptotically stable
and, therefore, the range scan sensor's location estimate $\bm{l}_t[k]$
converges to its real location $\bm{l}_t$ as $k \rightarrow \infty$. \qed

\subsection{Proof of convergence and stability when disturbances are present}
  \label{subsec:method_proofs_with_disturbance}
Suppose now that the ranges of the real scan $\mathcal{S}_r$ are corrupted
by additive, bounded disturbances $w_n^r$, where $0 < |w_n^r| \leq W_r$,
for all $n \in [0,N)$, and that those of the virtual scan are similarly
affected by $w_n^v$, $0 < |w_n^v| \leq W_v$ for all $n \in [0,N)$. Then,
denoting by $\widetilde{d}_n^r = d_n^r + w_n^r$ the range of ray $n$ measured
from the real range scan sensor, and by $\widetilde{d}_n^v = d_n^v + w_n^v$ the
range obtained from the virtual sensor, and applying the same rationale as in
subsection \ref{subsec:method_proofs_without_disturbance}, we arrive at the
analogous expression to (\ref{eq:final_errors_xy}):
\begin{align}
  \begin{bmatrix}
    e_x \\
    e_y
  \end{bmatrix} &= \dfrac{1}{N}
  \begin{bmatrix}
    \cos\theta & \sin\theta \\
    \sin\theta & - \cos\theta
  \end{bmatrix}
  \begin{bmatrix}
    \widetilde{X}_{1,r} \\
    \widetilde{X}_{1,i}
  \end{bmatrix}+
  \begin{bmatrix}
    \delta_x \\
    \delta_y
  \end{bmatrix}-
  \begin{bmatrix}
    w_x\\
    w_y
  \end{bmatrix} \nonumber \\
  &=
  \begin{bmatrix}
    \widetilde{u}_x\\
    \widetilde{u}_y
  \end{bmatrix}+
  \begin{bmatrix}
    \delta_x \\
    \delta_y
  \end{bmatrix}-
  \begin{bmatrix}
    w_x\\
    w_y
  \end{bmatrix} \nonumber
\end{align}
where
\begin{align}
  w_x &= \dfrac{1}{N} \sum\limits_{n=0}^{N-1} (w_n^r - w_n^v)\cos(\theta+\dfrac{2\pi n}{N}) \nonumber \\
  w_y &= \dfrac{1}{N} \sum\limits_{n=0}^{N-1} (w_n^r - w_n^v)\sin(\theta+\dfrac{2\pi n}{N}) \nonumber
\end{align}
$0 < |w_x| \leq W_x < \infty$, $0 < |w_y| \leq W_y < \infty$, i.e. $w_x$ and
$w_y$ are bounded since they are products of finite operations on finite
quantities, $\widetilde{X}_{1,r}$ and $\widetilde{X}_{1,i}$ are, respectively,
the real and imaginary parts of the complex quantity $\widetilde{X}_1$:
\begin{align}
  \widetilde{X}_1 &= \sum\limits_{n=0}^{N-1}(\widetilde{d}_n^r - \widetilde{d}_n^v) \cdot e^{-i \frac{2 \pi n}{N}} \nonumber
\end{align}
and
\begin{align}
  \begin{bmatrix}
    \widetilde{u}_x\\
    \widetilde{u}_y
  \end{bmatrix} &=
  \begin{bmatrix}
    u_x \\
    u_y
  \end{bmatrix}+
  \begin{bmatrix}
    w_x\\
    w_y
  \end{bmatrix} \label{eq:u_old_plus_w}
\end{align}
where $\bm{u} = [u_x, u_y]^{\top}$ is the control vector of the unperturbed
system (\ref{eq:proof_unperturbed_system_error}) and
$\widetilde{\bm{u}} = [\widetilde{u}_x, \widetilde{u}_y]^{\top}$ is the input
to the (now perturbed) system. Let us now denote the disturbance by
$\bm{w} = [w_x, w_y]^{\top}$; then the dynamics of the error of the perturbed
system become:
\begin{align}
  \bm{e}[k+1] &= \bm{e}[k] - \widetilde{\bm{u}}[k] \nonumber \\
              &\stackrel{(\ref{eq:u_old_plus_w})}{=} \bm{e}[k] - \bm{u}[k] - \bm{w}[k] \label{eq:proof_perturbed_system_error}
\end{align}

Since $\bm{w}$ is bounded, and the origin of the unperturbed system
(\ref{eq:proof_unperturbed_system_error}) is uniformly asymptotically stable,
the solution $\bm{e}[k]$ of the perturbed system
(\ref{eq:proof_perturbed_system_error}) is uniformly bounded for $k \geq k_0$
and uniformly ultimately bounded with an ultimate bound which is a function
of $\overline{W}_x$, $\overline{W}_y$, where $W_x \leq \overline{W}_x$,
$W_y \leq \overline{W}_y$ \citep{stability} (Theorem $2.7$, p.$29$). \qed

\subsection{In algorithmic form}
  \label{subsec:method_algorithmically}
The process of improving the estimate of the sensor's pose by the proposed
method---dubbed Iterative Correspondenceless Translation Estimator (ICTE)--- is
more clearly illustrated in pseudocode, where it is described in
Algorithm \ref{alg:algorithm_1}. The process' inputs are the map $M$ of the
robot's environment, the range scan captured from the true pose of the real
sensor $\mathcal{S}^t_r$ at time $t$, its properties, the estimated pose of the
sensor $\hat{\bm{p}}_t(\hat{x}_t, \hat{y}_t, \hat{\theta}_t)$, the threshold of
maximum iterations to run $k_{max}$, and the error threshold $\varepsilon_u$.
The map-scan $\mathcal{S}_v^{[k]}$ for the $k$-th iteration is computed in line
$4$ (Algorithm \ref{alg:algorithm_2}), and, together with the real scan, the
two are inputted to the \texttt{diffDFT} routine (Algorithm
\ref{alg:algorithm_3}). The latter's output is a complex number, $X_1$, whose
real and imaginary parts are used in lines $7$,$8$ to compute the $x$-wise and
$y$-wise corrections of the sensor's pose estimate (lines $9,10$). If the
$2$-norm of the control vector $\bm{u}_k = [u_x,u_y]^{\top}$ is found to be
below the preset threshold $\varepsilon_u$ (line $11$), the algorithm halts and
returns its last estimate; otherwise it iterates through the same sequence of
steps until this last condition is met, or until it runs out of iterations.

\begin{algorithm}
  \caption{\texttt{icte}}
  \begin{spacing}{1.2}
    \begin{algorithmic}[1]
      \REQUIRE M, $\mathcal{S}_r^t$, sensorProperties, $\hat{\bm{p}}_t(\hat{x}_t, \hat{y}_t, \hat{\theta}_t)$, $k_{max}$, $\varepsilon_u$
      \ENSURE $\hat{\bm{p}}^\prime_t(\hat{x}^{\prime}_t, \hat{y}^{\prime}_t, \hat{\theta}_t)$
      \STATE $k \leftarrow 0$
      \STATE $N = \text{sensorProperties}.N$
      \WHILE {$k < k_{max}$}
      \STATE $\mathcal{S}_v^{[k]} \leftarrow \text{scanMap}(M, \hat{\bm{p}}_t, \text{sensorProperties})$
      \STATE $X_1 \leftarrow \text{diffDFT}(\mathcal{S}_r^t, \mathcal{S}_v^{[k]})$
      \STATE $(X_{1,r}, X_{1,i}) \leftarrow (Re(X_1), Im(X_1))$
      \STATE $u_x \leftarrow (\cos\hat{\theta}_t \cdot X_{1,r} + \sin\hat{\theta}_t \cdot X_{1,i}) /N $
      \STATE $u_y \leftarrow (\sin\hat{\theta}_t \cdot X_{1,r} - \cos\hat{\theta}_t \cdot X_{1,i}) /N $
      \STATE $\bm{u}_k \leftarrow (u_x,u_y)$
      \STATE $\hat{\bm{p}}_t \leftarrow \hat{\bm{p}}_t + \bm{u}_k$
      \IF {$\|\bm{u}_k\|_2 < \varepsilon_u$}
        \STATE break
      \ENDIF
      \STATE $k \leftarrow k + 1$
      \ENDWHILE
      \STATE $\hat{\bm{p}}_t^\prime \leftarrow \hat{\bm{p}}_t$
      \RETURN $\hat{\bm{p}}_t^\prime$
    \end{algorithmic}
  \end{spacing}
  \label{alg:algorithm_1}
\end{algorithm}

Algorithms \ref{alg:algorithm_2} and \ref{alg:algorithm_3} respectively describe
the inner processes \texttt{scanMap}($\cdot$) and \texttt{diffDFT}($\cdot$) of
Algorithm \ref{alg:algorithm_1}.

In Algorithm \ref{alg:algorithm_2}, the routine for computing a map-scan is
outlined: in line $5$, $\lambda_n$ represents the angle that the scan's $n$-th
ray forms with respect to the sensor's $x$ axis in its local frame of reference,
and in line $6$ it is expressed with respect to the map's frame of reference
($\theta_n$). The intersection point of that ray, with a starting location that
of the sensor's pose estimate, and an orientation equal to $\theta_n$ and the
map is then computed at line $7$, and its range from the estimated pose of the
sensor is calculated in line $8$. This process is carried out sequentially
for each ray $n \in [0,N)$.

\begin{algorithm}
  \caption{\texttt{scanMap}}
  \begin{spacing}{1.2}
    \begin{algorithmic}[1]
      \REQUIRE M, $\hat{\bm{p}}_t(\hat{x}_t, \hat{y}_t, \hat{\theta}_t)$, sensorProperties
      \ENSURE $\mathcal{S}_v^{[k]}$
      \STATE $N = \text{sensorProperties}.N$
      \STATE $\lambda = \text{sensorProperties}.\lambda$
      \STATE $\mathcal{S}_v^{[k]} \leftarrow \{\}$
      \FOR{$n = 0:\text{sensorProperties}.N-1$}
      \STATE $\lambda_n \leftarrow - \dfrac{\lambda}{2} + n \dfrac{\lambda}{N} $
      \STATE $\theta_n \leftarrow \lambda_n + \hat{\theta}_t $
      \STATE $(x_n,y_n) \leftarrow \text{intersect}(M, (\hat{x}_t, \hat{y}_t, \theta_n))$
      \STATE $d_n^v \leftarrow \|(\hat{x}_t - x_n, \hat{y}_t - y_n)\|_2$
      \STATE Append $(d_n^v, \lambda_n)$ to $\mathcal{S}_v^{[k]}$
      \ENDFOR
      \RETURN $\mathcal{S}_v^{[k]}$
    \end{algorithmic}
  \end{spacing}
  \label{alg:algorithm_2}
\end{algorithm}

Algorithm \ref{alg:algorithm_3} illustrates the elementary routine for computing
the first term of the Discrete Fourier transform between ranges of homologous
rays of $\mathcal{S}_r^t$ and $\mathcal{S}_v^{[k]}$. $\bm{\Delta}$ is a
one-dimensional vector of equal size to the number of rays of each scan.
The $n$-th element of $\bm{\Delta}$ holds the range difference between the
$n$-th ray of $\mathcal{S}_r^t$ and that of $\mathcal{S}_v^{[k]}$. The
directive DFT($\bm{\Delta}$) in line $8$ computes the (complex) terms of the
Discrete Fourier transform of $\bm{\Delta}$ and, finally, in line $9$, the
first term (equation (\ref{eq:X1})) is extracted as the second element of the
returned vector (assuming zero-based indexing).

\begin{algorithm}
  \caption{\texttt{diffDFT}}
  \begin{spacing}{1.2}
    \begin{algorithmic}[1]
      \REQUIRE $\mathcal{S}_r^t$, $\mathcal{S}_v^{[k]}$
      \ENSURE $X_1$
      \STATE Assert that size of $\mathcal{S}_r^t$ equals size of $\mathcal{S}_{v}^{[k]}$
      \STATE $N \leftarrow$ size of $\mathcal{S}_r^t$
      \STATE {$\bm{\Delta} \leftarrow \{\}$}
      \FOR{$n = 0:N-1$}
      \STATE $d \leftarrow \mathcal{S}_{r}^{t}.d_n - \mathcal{S}_{v}^{[k]}.d_n$
      \STATE Append $d$ to $\bm{\Delta}$
      \ENDFOR
      \STATE $\bm{X} \leftarrow \text{DFT}(\bm{\Delta})$
      \STATE $X_1 \leftarrow \bm{X}[1]$
      \RETURN $X_1$
    \end{algorithmic}
  \end{spacing}
  \label{alg:algorithm_3}
\end{algorithm}

\subsection{Commentary}
  \label{subsec:method_commentary}
The convergence of the location estimate and its final error depend mainly on
the disturbance acting on the readings of the real and the virtual range sensor.
With respect to the former, range inaccuracy with respect to real distances to
obstacles in the sensor's environment is not only due to the real sensor's
finite range resolution, but depends also on the material of the target surface,
its distance with respect to the obstacle, the sensor's temperature, and others
\citep{lidar}. As for the virtual scan sensor, we refer to acting disturbances
in the sense of map inaccuracies: when the map does not accurately correspond
to the environment, ranges captured from within the map are inaccurate with
respect to the environment and, therefore, subject to perceived noise.

\begin{figure}[!htb]\centering
  \input{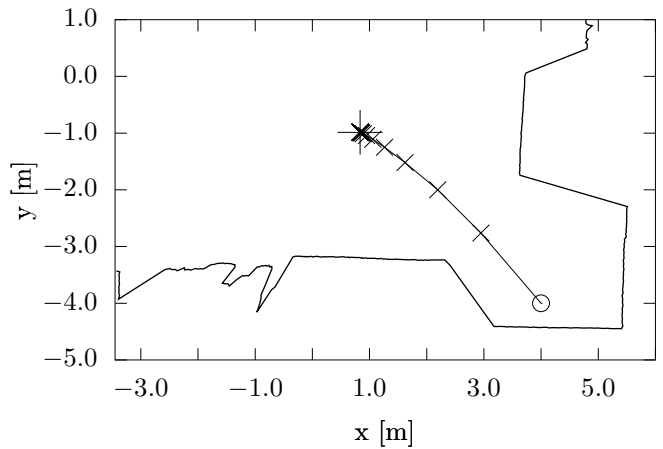}
  \vspace{0.3cm}
  \caption{Estimate correction illustration in the absence of disturbances: the
           map captures the environment perfectly and sensor noise is absent.
           The true location of the robot is depicted with a cross, the initial
           estimate with a circle, and the intermediate location estimates with
           $\times$ marks}
  \label{fig:map_perfect}
\end{figure}

\begin{figure}[!htb]\centering
  \hbox{\hspace{-0.5cm}\input{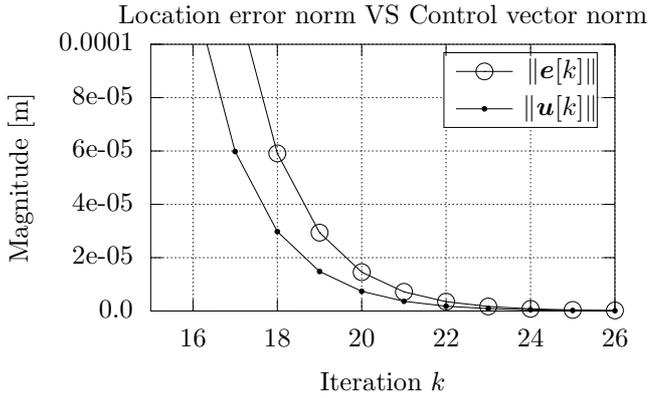}}
  \caption{Estimate correction illustration in the absence of disturbances.
           Dots signify the norm of the control vector at each iteration.
           Circles signify the norm of the true location error at each
           iteration. Both curves are strictly decreasing}
  \label{fig:convergence_perfect}
\end{figure}

In the absence of disturbances, ranges from both sensors capture the environment
perfectly and the sensor's (robot's) location estimate can be brought
arbitrarily close to the true location of the sensor (robot). Figure
\ref{fig:map_perfect} illustrates the evolution of the location estimate
within a sample map, while figure \ref{fig:convergence_perfect} illustrates the
evolution of the norms of the location error $\bm{e} = [e_x, e_y]^{\top}$ and
the control vector $\bm{u} = [u_x, u_y]^{\top}$ as a function of time. In this
case, convergence is asymptotic, and the location estimate error is driven
exclusively by the control vector.

\begin{figure}[!htb]
  \hbox{\hspace{-0.5em}
  \input{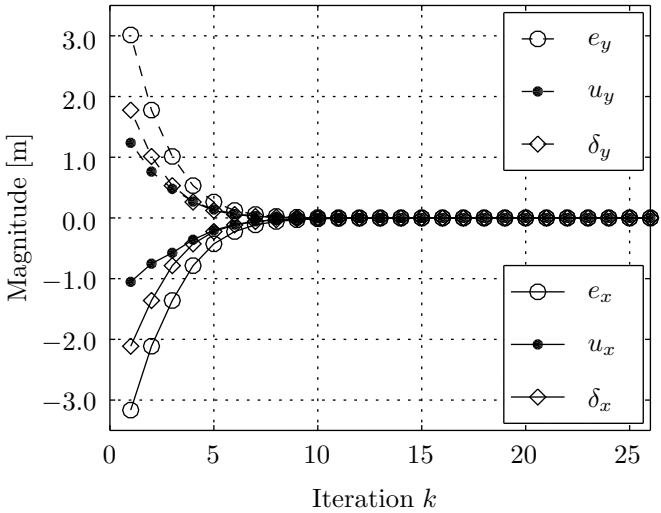}}
  \vspace{-7.9cm}
  \caption{Estimate correction illustration in the absence of disturbances.
           Circles signify the value of the error component, dots the value of
           the control vector component, and diamonds the value of the component
           of function $\delta$}
  \label{fig:data_perfect}
\end{figure}

\begin{figure}[!htb]\centering
  \input{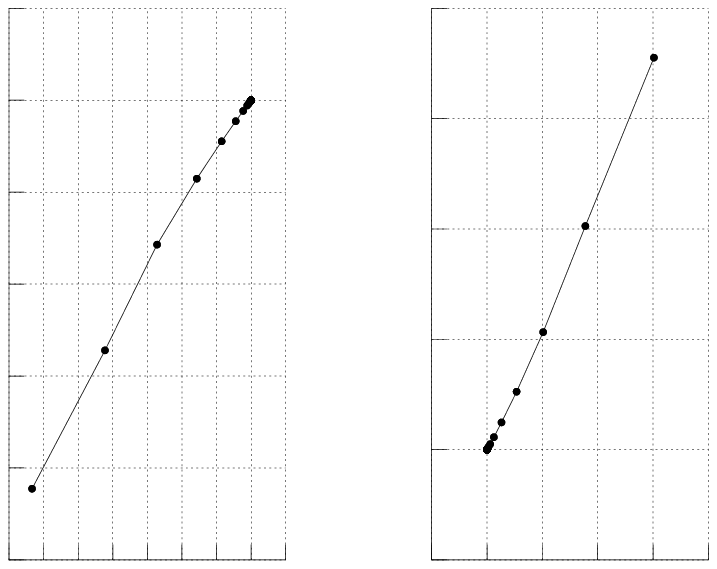}
  \vspace{0.3cm}
  \caption{The evolution of function $\bm{\delta}$ components against their
           corresponding error component in the absence of disturbances. Both
           curves converge to zero as time grows and both are strictly
           increasing with respect to increasing error}
  \label{fig:delta_perfect}
\end{figure}

Figure \ref{fig:data_perfect} illustrates
the evolution of the error $\bm{e}$, control vector $\bm{u}$, and function
$\bm{\delta} = [\delta_x, \delta_y]^{\top}$ per-axis components as a function
of time, all converging to zero as time grows. Figure \ref{fig:delta_perfect}
depicts the components of $\bm{\delta}$ as a function of the error $\bm{e}$ in
the corresponding axis; both functions are strictly increasing.

\begin{figure}[!htb]\centering
  \hbox{\hspace{0.15em}
  \input{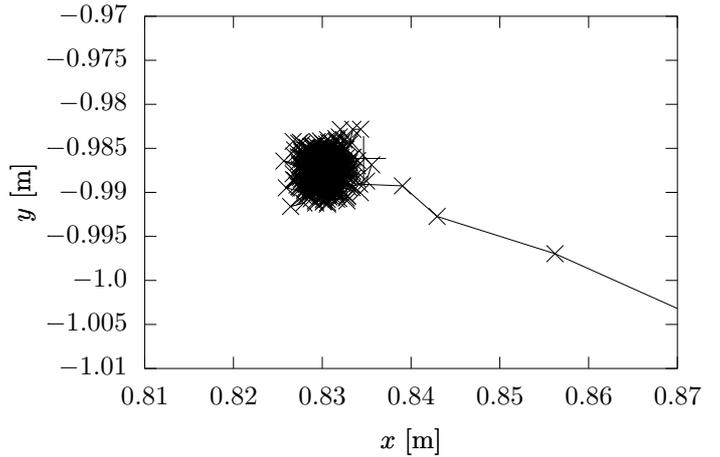}}
  \vspace{0.3cm}
  \caption{Estimate correction illustration in the presence of disturbances: the
           the real scan and the virtual scans are perturbed by zero-mean,
           normally-distributed noise with standard deviation equal to
           $\sigma = 0.05$ m. The true location of the robot is depicted with a
           cross and the intermediate location estimates with $\times$ marks. The
           location error is guaranteed to be bound, therefore the location
           estimate trajectories are bound to lie in a neighbourhood of the
           true location}
  \label{fig:convergence_imperfect_500}
\end{figure}

\begin{figure}[!htb]\centering
  \hbox{\hspace{-0.5cm}
  \input{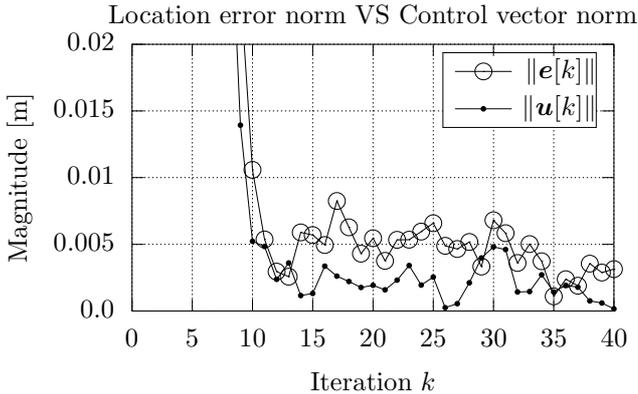}}
  \caption{Estimate correction illustration in the presence of disturbances: the
           the real scan and the virtual scans are perturbed by zero-mean,
           normally-distributed noise with standard deviation equal to
           $\sigma = 0.05$ m. Dots signify the norm of the control vector at
           each iteration. Circles signify the norm of the location error at
           each iteration. Note how the minimum of the location error's norm
           does not correspond to that of the control vector anymore}
  \label{fig:convergence_imperfect}
\end{figure}

By contrast, when noise is present in the ranges of either the real or the
virtual range scan sensor, the control vector does not unilaterally determine
the estimate error. Figure \ref{fig:convergence_imperfect_500} depicts the
trajectory of the location estimate near the target location when both real
and virtual scans are perturbed by zero-mean, normally-distributed noise with
standard deviation equal to $\sigma = 0.05$ m, given the same map and initial
location as in the case of absent disturbances. Due to the presence of noise,
the location estimate is offset and unable to converge arbitrarily close to it.
Here, $\varepsilon_u$, the error threshold for stopping, has been set high
enough so that the algorithm terminates only due to reaching the maximum number
of iterations, set here to $k_{max} = 600$. Figure
\ref{fig:convergence_imperfect} illustrates the evolution of the norms of the
location error $\bm{e}$ and the control vector $\bm{u}$ as time progresses,
where it is evident that both are not strictly decreasing from some iteration
forward (for $k \geq k_0 = 13$ in particular). In parallel, for all
$k \geq k_0$, the error is bound, as guaranteed by Theorem
\ref{prop:theorem_with_disturbance}; the trajectory of the estimate location
(figure \ref{fig:convergence_imperfect_500}) does not diverge for any
$k \in [0,k_{max}]$.

Finally, the proposed method is akin to the update step of the (Extended)
Kalman filter: both assume as inputs a prior pose estimate, a measurement
vector at the current time step, and the map of the environment, and both
attempt to improve the accuracy of this prior estimate by utilising a measure
of divergence of measurements from their estimated values (compare eq.
(\ref{eq:X1}) to the calculation of the innovation in the update step of the
Kalman filter). In the Kalman filter's case this prior is issued from the
robot's motion model, but in the case of the proposed method this prior may be
issued from additional sources. However, their differences lie in the facts that
the proposed method (a) assumes that the orientation of the robot is known, and
(b) does not make assumptions on the distribution of noise affecting the range
measurements of either the physical or the virtual range sensor, apart from
it being additive and bounded.

\section{Experimental Procedure}
  \label{section:experimental}

This section serves to illustrate the efficacy and performance of the proposed
method. The following experiments were conducted using a benchmark dataset
consisting of $778$ laser scans obtained from a Sick range-scan sensor mounted
on a robotic wheel-chair emitting 360 rays over a $180^\circ$ field of
view\footnote{\url{https://censi.science/pub/research/2007-plicp/laserazosSM3.log.gz}}.
The same dataset was used to evaluate the performance of IDC \citep{idc}, ICP,
and MBICP in \citep{mbicp}, and that of PLICP in \citep{plicp}, wherein the latter
was found to be the best-performing among the four correspondence-finding
state-of-the-art scan-matching methods. For each scan, the dataset reports
$360$ range measurements and a pose $\bm{p}(x,y,\theta)$ from which it was
captured.

The conducted experiments test for performance in two regards and with two
discrete objectives: (a) to query on the difference of mean error between the
proposed method, the optimal correspondence-finding scan-matching method
(PLICP), and the equally correspondenceless but probabilistic scan-matching
method of NDT when employed in a scan--to--map-scan matching context, and (b) to
query on how the proposed method's mean error varies with respect to common and
varying limitations of the range-scan sensor, such as varying field of view or
number of rays emitted, and real-life occurrences of failure, such as retrieval
of rays with invalid range or the reasonable non-perfection of the method that
aligns the sensor's orientation estimate with the sensor's true orientation,
typically needed before the proposed method can be employed. The results of
tests relative to the first category are found in subsection
\ref{subsec:fsm_csm_ultimate_showdown}, while those relative to the second in
subsection \ref{subsec:with_limitations}.

In the following we describe how the inputs of the proposed method (which are
the same as those of PLICP and NDT) are constructed given only a dataset
instance.

The proposed method requires as inputs: (a) a map $M$ of the robot's
surroundings, (b) the sensor's pose estimate $\hat{\bm{p}}$ within $M$,
(c) a range scan $\mathcal{S}_r$ ranging over $2\pi$, (d) an upper threshold of
iterations $k_{max}$, and (e) a threshold for stopping $\varepsilon_u$.

As the proposed method requires a map, dataset instances were used to construct
a map as follows. Let a dataset $D$ instance $d$ comprising $N_d = 360$
range measurements $\bm{R}_d = \{r^i_d\}$,
$r^i_d \in \mathbb{R}$, $i = \{0,1,\dots N_d-1\}$ be captured from pose
$\bm{p}_d \equiv (px^i_d, py^i_d, p\theta^i_d) \in \mathbb{R}^2 \times [-\pi,\pi)$.
A map $M_d^0$ corresponding to the $d$-th scan, $d = \{0,1,\dots,N_D - 1\}$,
$N_D = 778$, is constructed as a collection of points, where the coordinates of
each point in the $x-y$ plane are respectively $
x^i_d = px^i_d + r^i_d \cos(-\pi/2 + i\pi/ N_d + p\theta^i_d)$ and
$y^i_d = py^i_d + r^i_d \sin(-\pi/2 + i\pi/ N_d + p\theta^i_d)$. Since $M_d^0$
ranges over a field of view of $\pi < 2\pi$ rads for the particular dataset,
there is a range of options on how to supplement it so that it ranges over
an angular field of view of $2\pi$ and bridge the gap non-arbitrarily that we
identify: (a) mirror the points of $M_d^0$ with respect to $\bm{p}_d$ with the
axis of symmetry set to that corresponding to the $x$ axis of the robot
(assuming the right-handed 3D coordinate frame convention) (b) mirror them with
the axis of symmetry that of the $y$ axis of the robot, and (c) draw a
semicircular arc around $\bm{p}_d$ with radius set to the minimum range between
the two extreme rays. All three have been found equivalent with respect to the
method's performance.

The initial estimated pose of the virtual sensor $\hat{\bm{p}}$ is obtained for
each dataset instance by perturbing the $x$ and $y$ axis components of
$\bm{p}_d$ with quantities extracted from a uniformly distributed
error distribution $U(-\alpha, \alpha)$, $\alpha \in \mathbb{R}$.

A virtual scan $\mathcal{S}_v$ is then produced by locating the intersections
of rays emanating from the virtual sensor's estimated pose $\hat{\bm{p}}$ and
ranging over $\lambda = 2\pi$ within the produced map $M_d$ with the lines
connecting its points. The corresponding real laser scan $\mathcal{S}_r$ is
obtained by augmenting vector $\bm{R}_d$ by appending it with the range of rays
captured from $\bm{p}_d$ within $M_d$ from the missing
$[\pi/2, \pi) \cup [-\pi, -\pi/2)$ range. Lastly, the iterations threshold was
set to $k_{max} = 60$, and the threshold for stopping to
$\varepsilon_u = 10^{-5}$ m.

In order to test for the performance of the proposed method, we test for four
discrete values of $\alpha$, in order to progressively test it in the range of
pose estimate error values typically reported in the bibliography:
$\alpha \in \{0.05, 0.10, 0.15, 0.20\}$ m.

Furthermore, we test for five different levels of disturbances acting on the
range measurements of the real and virtual scans, as these may manifest
themselves in real conditions: Range scans are affected by additive zero-mean
normally distributed noise with standard deviation equal to
$\sigma \in \{0.0, 0.005, 0.01, 0.02, 0.05\}$ m. For each displacement value,
each noise level, and each case, we run Algorithm \ref{alg:algorithm_1} for $I
= 100$ times for each dataset, totaling $4 \times 5^2 \times 6 \times 100
\times 788 \sim \mathcal{O}(10^7)$ runs.  PLICP and NDT were ran only for the
nominal case, i.e. without querying their performance under limitations or
failures, for a total of $4 \times 5^2 \times 100 \times 778 \sim
\mathcal{O}(10^6)$ runs. All experiments and all algorithms ran serially, on a
single thread, in a machine with a CPU frequency of $4.00$ GHz.

Unless otherwise noted, in the following figures circles ($\bigcirc$) denote
the performance of either the proposed method, PLICP, or NDT when $\alpha =
0.05$ m, stars ($\ast$) when $\alpha = 0.10$ m, downward-facing triangles
($\triangledown$) when $\alpha = 0.15$ m, and squares ($\square$) when $\alpha
= 0.20$ m.

\subsection{Comparison against the state of the art}
  \label{subsec:fsm_csm_ultimate_showdown}
\begin{figure}\hspace{1.5cm}
  \input{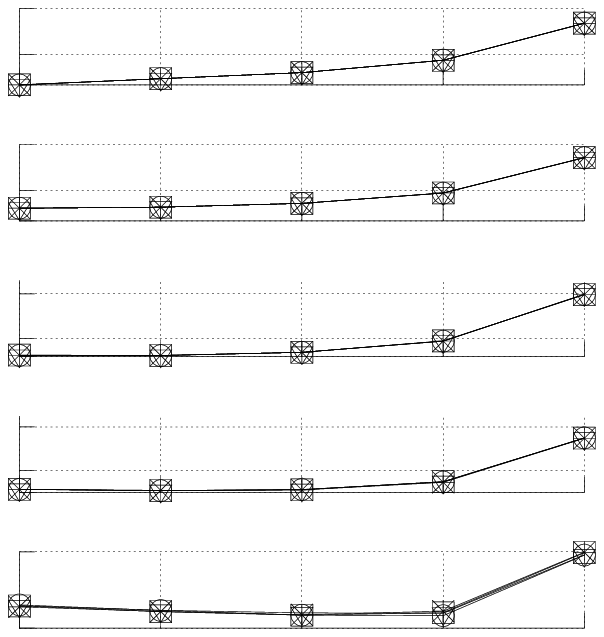}
  \vspace{0.2cm}
  \caption{Mean position errors of the proposed method over $I = 100$ runs for
           each combination of real scan noise standard deviation $\sigma_r$,
           virtual scan standard deviation $\sigma_v$, and displacement $\alpha$ }
  \label{fig:performance_proposed_method}
\end{figure}

\begin{figure}\hspace{1.5cm}
  \input{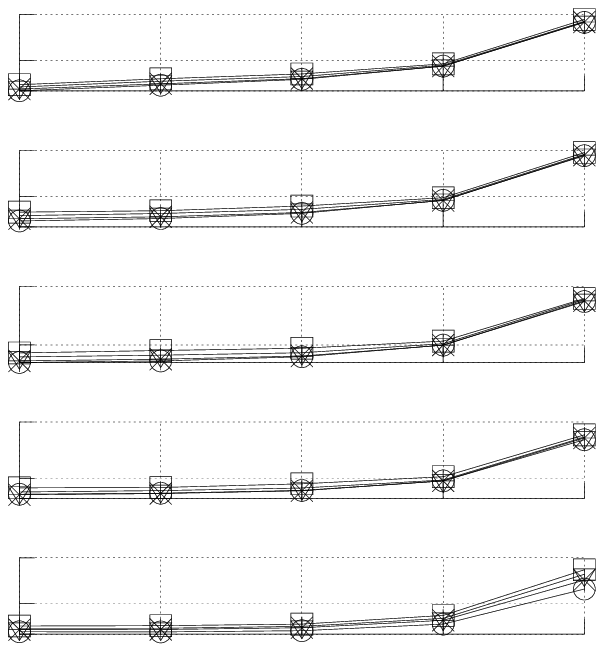}
  \vspace{0.2cm}
  \caption{Mean position errors of PLICP over $I = 100$ runs for each
           combination of real scan noise standard deviation $\sigma_r$, virtual
           scan standard deviation $\sigma_v$, and displacement $\alpha$ }
  \label{fig:performance_PLICP}
\end{figure}

\begin{figure}[]\hspace{1.5cm}
  \input{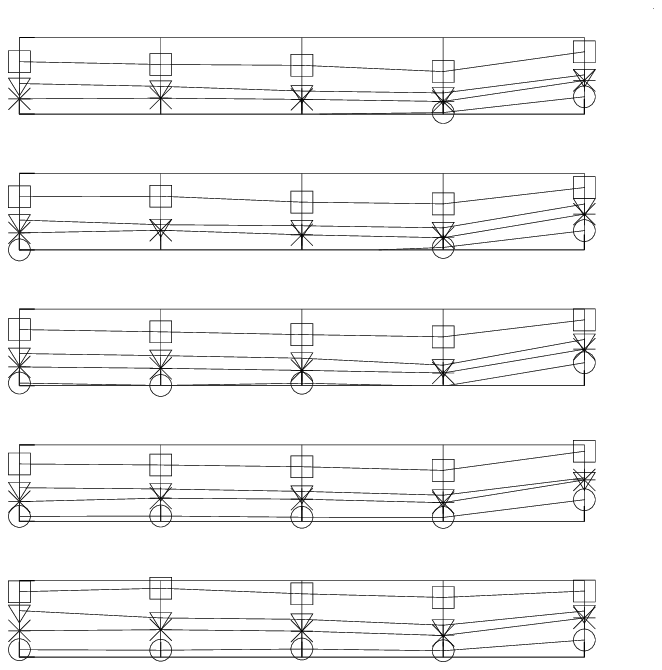}
  \vspace{0.2cm}
  \caption{Mean position errors of NDT over $I = 100$ runs for each
         combination of real scan noise standard deviation $\sigma_r$, virtual
         scan standard deviation $\sigma_v$, and displacement $\alpha$}
  \label{fig:performance_NDT}
\end{figure}

Figure \ref{fig:performance_proposed_method} illustrates the mean error of the
proposed method on the tested dataset for different levels of sensor position
displacement, real scan noise levels, and virtual scan noise levels, over
$I = 100$ runs. Figures \ref{fig:performance_PLICP} and \ref{fig:performance_NDT}
illustrate the mean error of the PLICP and NDT methods over the same dataset
for the same levels of displacement and scans' noise and over the same number
of runs.

\begin{figure}[]\hspace{1.5cm}
  \input{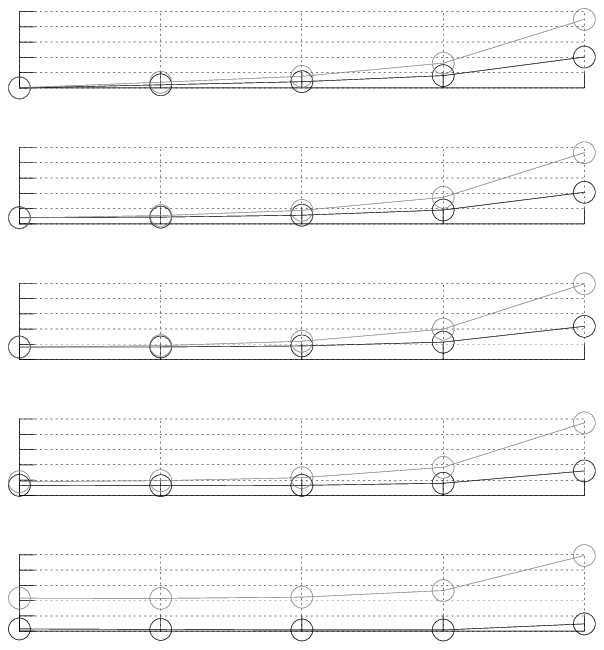}
    \vspace{0.2cm}
    \caption{Direct comparison of the mean errors of the proposed method (solid
             black) and PLICP (faint black) over all tested scan noise levels,
             for initial sensor position error $\alpha = 0.05$ m per $x,y$
             pose component}
    \label{fig:direct_comparison_5}
\end{figure}

The proposed method's mean errors are consistently invariant across different
levels of position displacement, i.e. in a neighbourhood of the true sensor
position, in terms of the position error, its performance does not depend on
the sensor's initial position estimate. With regard to PLICP, the proposed
method's mean errors are lower; their difference increases the more noise is
present in either real or virtual scan for the same level of position
displacement, or the greater the initial position error is for the same levels
of scans' noises. Figure \ref{fig:direct_comparison_5} depicts a direct
comparison between the proposed method and PLICP when $\alpha = 0.05$ m. With
regard to NDT, the proposed method's mean errors are significantly lower. NDT's
performance is also dependent on the initial displacement, but it demonstrates
a higher degree of robustness compared to PLICP, as its mean position errors
are almost invariant to either sensor noise or map-to-environment discrepancy
for a given level of initial displacement.



\begin{figure}\hspace{1.5cm}
  \input{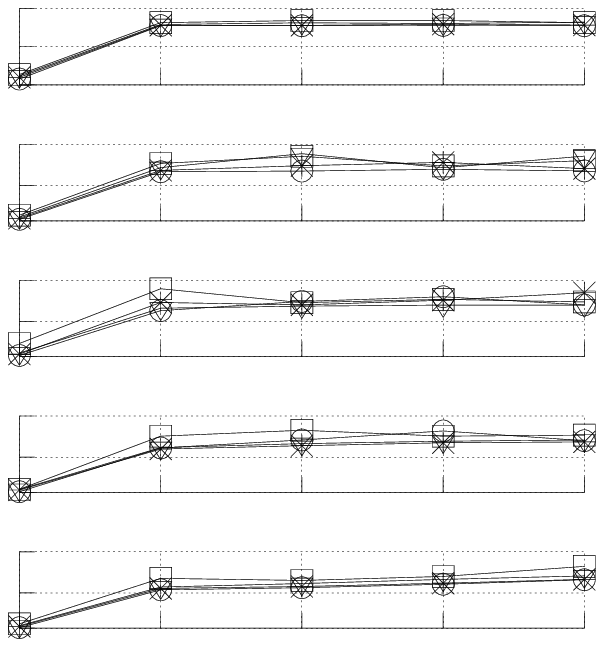}
    \vspace{0.2cm}
    \caption{Mean execution times corresponding to the experiments that produced
             figure \ref{fig:performance_proposed_method}}
    \label{fig:proposed_method_times}
\end{figure}

Figure \ref{fig:proposed_method_times} illustrates the corresponding mean
execution times of the proposed method. Since, in general, what ultimately
determines the length of execution is the combination of the maximum iterations
threshold $k_{max}$ and the threshold for stopping $\varepsilon_u$,
it is evident from figures \ref{fig:performance_proposed_method} and
\ref{fig:proposed_method_times} that the configuration
$(k_{max}, \varepsilon_u) \equiv (60, 10^{-5})$ is able to provide improved
position estimation at real-time for realistic levels of estimation error and
noise in scans. The longest overall execution did not take more than $25$ ms,
and therefore the highest frequency at which the method can be run is at
$40$ scan-matchings per second. PLICP's mean execution time varied between
$4$ ms in the case of minimal position estimate displacement and noise levels
and $28$ ms in the case of their maximal values. The corresponding mean
execution times for NDT were $195$-$307$ ms.

\subsection{A characterisation over limitations posed by reality}
  \label{subsec:with_limitations}
Commercially available LIDAR sensors vary in terms their field of view $\lambda$
and number of rays they emit $N$. In terms of the latter, the position error
exhibited by the proposed method decreases in proportion to the number of
available rays. This result stands to reason as, the denser the sampling of the
environment and the map, the better a virtual scan may approximate the real
scan, and therefore the closer the robot's position estimate may approximate
its actual position. In terms of a sensor's field of view, the proposed
method's location errors increase when the field of view is $\lambda < 2\pi$.
In that case location errors become proportional to the initial displacement
between the sensor's location estimate and its true location. Furthermore, the
proposed method reports reduced errors for a sensor with a field of view of
$\lambda = 3\pi/2$ in comparison to a panoramic sensor which emits $360$ rays
or fewer.

In real conditions, the range reported for a number of rays may be invalid
due to sensor fault. The proposed method reports a slight increase in
position errors for a percentage of randomly invalid rays in the range of
$10\%$-$20\%$ compared to the case of all rays being valid. Random invalidation
of half of the total rays emitted produces slight increase in position errors
compared to those reported for a sensor emitting all-valid rays but half in
number. This fact suggests that even sampling of space fares better than uneven
sampling.

Another limitation that manifests in real conditions is that, due to the
existence of a maximum detectable range by a LIDAR sensor, the range reported
for a number of consecutive rays may peak at its maximum range (e.g. when a
robot is placed at one end of a long corridor), thus reporting a
misrepresentation for some portion of the environment. As expected, the larger
these portions are in size, the greater the position error that the proposed
method exhibits. However, the results also show that for increasing position
displacements, the proposed method performs better when invalid rays are
concentrated over one region compared to their being scattered unevenly.

Finally, the orientation estimate provided to the proposed method may be
accurate to a certain extent. Simulations show that for the range of
orientation estimate errors reported in the bibliography, the proposed method's
position errors may increase by as much as five times compared to the case of
complete orientation coincidence, if these orientation errors are left
untreated.

The complete results for the above cases are reported in the appendix, in
figures \ref{fig:undersample}-\ref{fig:rotation}.

\section{Conclusions}
  \label{section:finale}
The premise of this study was improving the location estimate of a mobile robot
capable of motion on a plane and mounted with a conventional 2D LIDAR sensor,
given an initial guess for its location on a 2D map of its surroundings.
The preceding analysis provided the theoretical reasoning behind solving a
matching problem between two homoriented 2D scans, one derived from the robot's
physical sensor and one derived by simulating its operation within the map,
in a way that does not require the establishing of correspondences between their
constituting rays. Two results were proved and subsequently shown through
experiments. The first is that the true position of the sensor can be
recovered with arbitrary precision when the physical sensor reports faultless
measurements and there is no discrepancy between the environment the robot
operates in and its perception of it by the robot. The second is that when
either is affected by disturbance, the location estimate is bound in a
neighbourhood of the true location whose radius is proportional to the
affecting disturbance.


\newpage
\cleardoublepage

\section{Appendix}
  \label{section:appendix}

This section houses the results of simulations of the proposed method which are
conducted over a range of constraints posed by real conditions
(section \ref{subsec:with_limitations}).

\begin{figure}[H]
  \hbox{\hspace{4.5em}
  \input{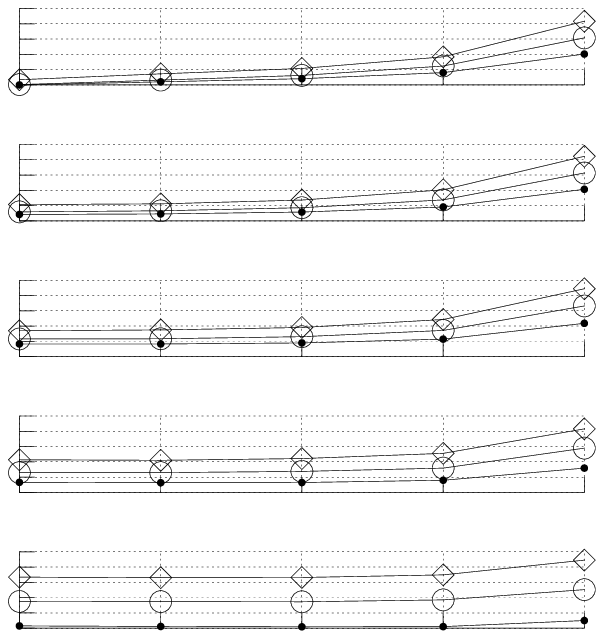}}
  \vspace{0.5cm}
  \caption{Mean position errors of the proposed method over $I = 100$ runs
           for each combination of real scan noise standard deviation
           $\sigma_r$, virtual scan standard deviation $\sigma_v$, for
           displacement $\alpha = 0.2$ m, for different number of rays emitted
           from the range sensor. Dots signify the nominal configuration ($720$
           rays), circles a reduction by half ($360$ rays), and diamonds
           by a third ($240$ rays)}
  \label{fig:undersample}
\end{figure}

Figure \ref{fig:undersample} depicts the proposed method's mean position errors
for varying number of sensor rays: nominal ($720$), half ($360$), and one
third ($240$) when $\alpha = 0.2$ m (in fact all displacement configurations
yield the same results, as in the nominal case). Evidently, the error decreases
in proportion to the number of available rays.

\begin{figure}
  \hbox{\hspace{4em}
  \input{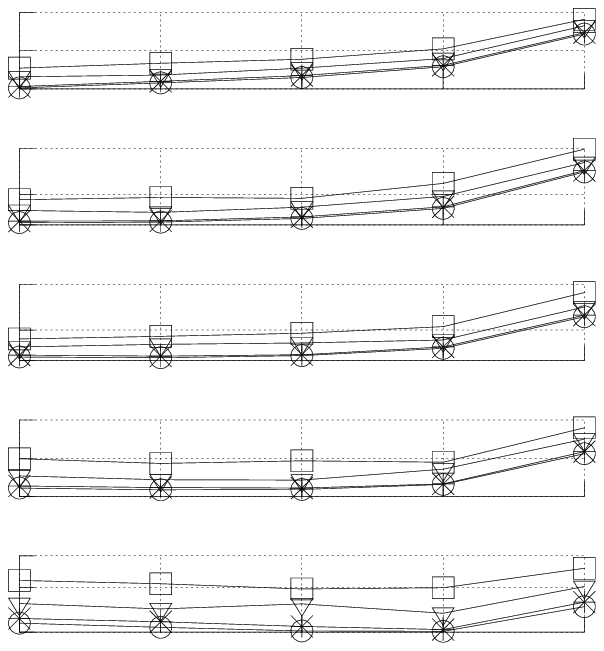}}
  \vspace{0.5cm}
  \caption{Mean position errors of the proposed method over $I = 100$ runs
           for each combination of real scan noise standard deviation
           $\sigma_r$, virtual scan standard deviation $\sigma_v$ and
           displacement $\alpha$ when the range sensor's field of view is
           reduced to $\lambda = 3\pi/2$ rad}
  \label{fig:dft_270}
\end{figure}

Figure \ref{fig:dft_270} illustrates the proposed method's mean position
errors when the range-finder sensor's field of view is
$\lambda = 3\pi/2 < 2\pi$ rad, distributed evenly over the sensor's $z$ axis.
Evidently, what determines the independence of the method's performance from
the initial location error is whether or not the range sensor has a panoramic
field of view.

\begin{figure}
  \hbox{\hspace{4em}
  \input{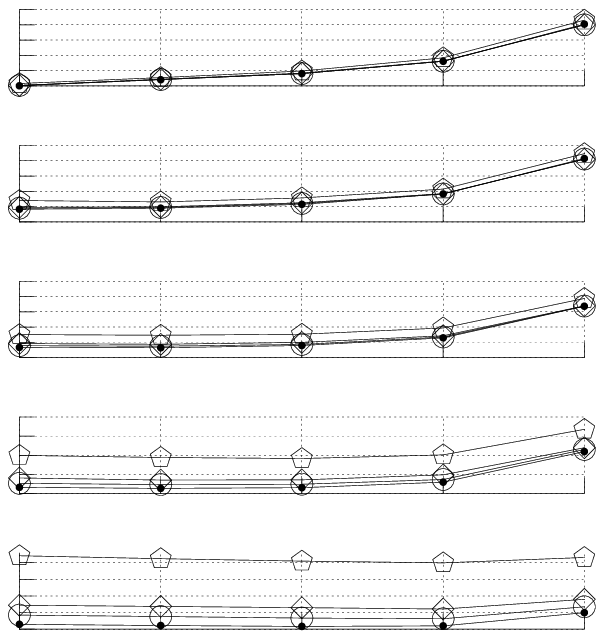}}
  \vspace{0.5cm}
  \caption{Mean position errors of the proposed method over $I = 100$ runs
           for each combination of real scan noise standard deviation
           $\sigma_r$, virtual scan standard deviation $\sigma_v$, for
           displacement $\alpha = 0.2$ m, for different levels of randomly
           invalidated number of rays: dots denote the nominal configuration
           (none invalid), circles denote a configuration where $10\%$ of
           all rays are randomly invalidated, diamonds when $20\%$ are
           invalidated, and pentagrams when $50\%$ are invalidated}
  \label{fig:invalid_randomly}
\end{figure}

Figure \ref{fig:invalid_randomly} depicts the proposed method's mean position
error for varying levels of randomly invalidated rays---scenaria of rather
uncommon failures ($10\% - 20\%$ of the nominal number of rays, which is $720$)
and extreme failures ($50\%$). When a ray is detected as invalid, the value of
its range along with the corresponding one from the map-scan is zeroed out but
included in the computation of the $X_1$ term (equation (\ref{eq:X1})).
For up to uncommon levels of sensor failure to retrieve range, the proposed
method's performance is relatively unaffected.

\begin{figure}
  \hbox{\hspace{5em}
  \input{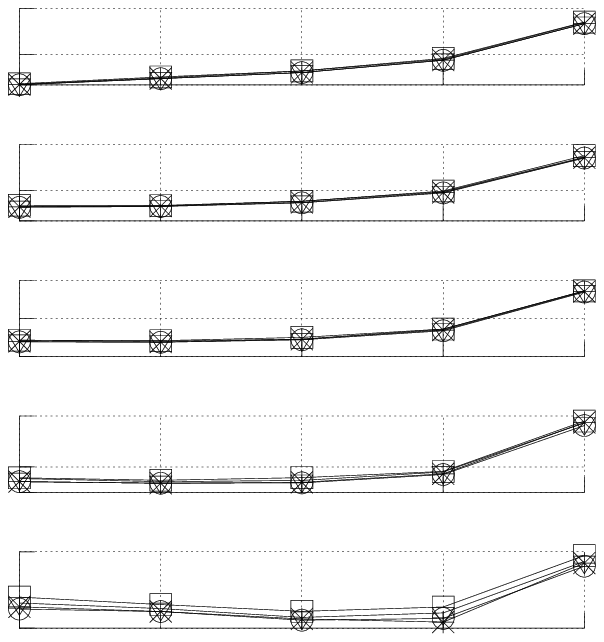}}
  \vspace{0.5cm}
  \caption{Mean position errors of the proposed method over $I = 100$ runs
           for each combination of real scan noise standard deviation
           $\sigma_r$, virtual scan standard deviation $\sigma_v$, and
           displacement $\alpha$, when $10\%$ of the sensor's rays
           are consecutively invalid}
  \label{fig:invalid_sequentially_01}
\end{figure}

\begin{figure}
  \hbox{\hspace{5em}
  \input{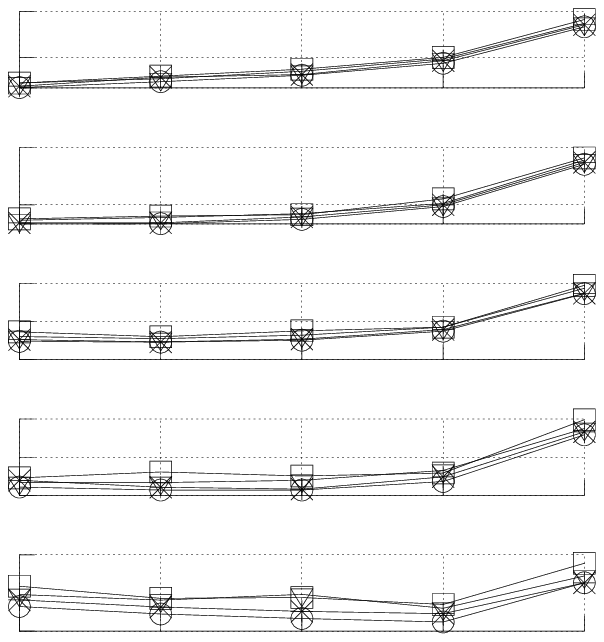}}
  \vspace{0.5cm}
  \caption{Mean position errors of the proposed method over $I = 100$ runs
           for each combination of real scan noise standard deviation
           $\sigma_r$, virtual scan standard deviation $\sigma_v$, and
           displacement $\alpha$, when $20\%$ of the sensor's rays
           are consecutively invalid}
  \label{fig:invalid_sequentially_02}
\end{figure}

\begin{figure}
  \hbox{\hspace{4em}
  \input{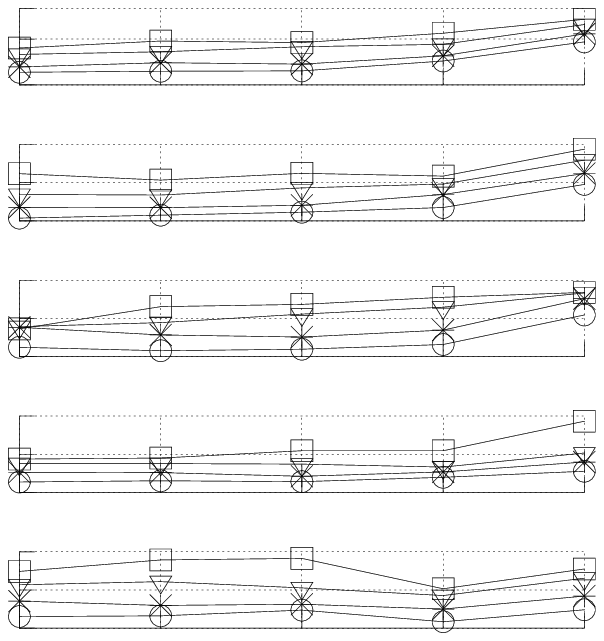}}
  \vspace{0.5cm}
  \caption{Mean position errors of the proposed method over $I = 100$ runs
           for each combination of real scan noise standard deviation
           $\sigma_r$, virtual scan standard deviation $\sigma_v$, and
           displacement $\alpha$, when $50\%$ of the sensor's rays
           are consecutively invalid}
  \label{fig:invalid_sequentially_05}
\end{figure}

Figures \ref{fig:invalid_sequentially_01}-\ref{fig:invalid_sequentially_05}
illustrate the proposed method's mean position errors for varying levels of
consecutively invalidated rays---common cases where obstacles are farther
away from the sensor than its maximum range. The index of the first ray from
which an invalid block of rays is established is chosen at random among all
rays. Analogously to the case where the index of an invalid range is chosen at
random between all, the performance of the proposed method is not significantly
affected when invalid blocks are relatively small in size, but deteriorates to
twice the mean error compared to the nominal case when half of all rays are
consecutively invalid.

\begin{figure}
  \hbox{\hspace{4em}
  \input{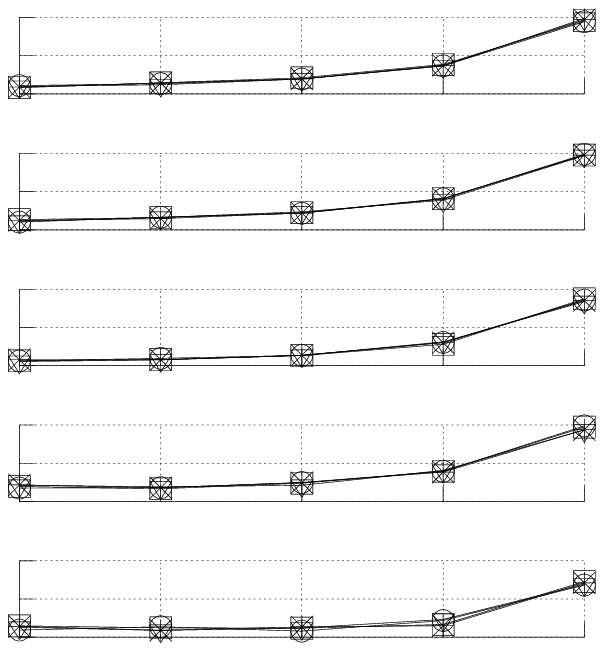}}
  \vspace{0.5cm}
  \caption{Mean position errors of the proposed method over $I = 100$ runs
           for each combination of real scan noise standard deviation
           $\sigma_r$, virtual scan standard deviation $\sigma_v$, and
           displacement $\alpha$, when the estimated pose of the sensor is
           rotationally misaligned with respect to its true pose. The rotation
           error is chosen at random from a uniform distribution
           $U_r = U(-\rho_1, -\rho_0) \cup U(\rho_0, \rho_1)$, $\rho_0 = 0.003$
           rad and $\rho_1 = 0.01$ rad}
  \label{fig:rotation}
\end{figure}

Figure \ref{fig:rotation} depicts the proposed method's mean errors when the
sensor's estimate is rotationally misaligned with regard to the sensor's true
pose. The rotational displacement for each experiment was chosen from the
uniform distribution $U_r = U(-\rho_1, -\rho_0) \cup U(\rho_0, \rho_1)$,
where $\rho_0 = 0.003$ rad and $\rho_1 = 0.01$ rad. The values of $\rho_{\ast}$
were selected from the high end of the spectrum of mean rotational errors
reported in the literature of scan--to--map-scan matching, as documented in
subsection \ref{subsec:sota_solutions}. Overall, from figure \ref{fig:rotation}
it is quite obvious that the performance of the proposed method relies heavily
on the precision of its antecedent method that angularly aligns the two scans.\\

\end{document}